\documentclass[final,5p,times,twocolumn]{elsarticle}

\usepackage{graphicx}
\usepackage[utf8]{inputenc}
\usepackage{graphics, subfig} 
\usepackage{epsfig} 
\usepackage{mathptmx} 
\usepackage{amsmath} 
\usepackage{amssymb}  
\usepackage{lineno,hyperref}
\modulolinenumbers[5]
\usepackage{subfig} 
\usepackage{bm} 
\usepackage[euler]{textgreek} 
\usepackage{mathtools}
\DeclarePairedDelimiterX{\norm}[1]{\lVert}{\rVert}{#1}
\usepackage[table]{xcolor}
\usepackage{multirow}
\usepackage{enumitem}
\usepackage{soul} 
\usepackage[ruled, linesnumbered]{algorithm2e} 
\usepackage{threeparttable} 

\journal{Information Fusion}

\newcommand{\ourMethod}{Light Federated and Continual Consensus}
\newcommand{\ourAcronym}{LFedCon2}
\begin{document}

\begin{frontmatter}

\title{Federated and continual learning for classification tasks in a society of devices}

\author[citius]{Fernando E. Casado\corref{mycorrespondingauthor}}
\cortext[mycorrespondingauthor]{Corresponding author}
\ead{fernando.estevez.casado@usc.es}

\author[citius]{Dylan Lema}\ead{dylan.lema@usc.es}

\author[citius]{Roberto Iglesias}\ead{roberto.iglesias.rodriguez@usc.es}

\author[fic]{Carlos V. Regueiro}\ead{carlos.vazquez.regueiro@udc.es}

\author[citius]{Sen\'en Barro}\ead{senen.barro@usc.es}

\address[citius]{CiTIUS (Centro Singular de Investigaci\'on en Tecnolox\'ias Intelixentes), Universidade de Santiago de Compostela, 15782 Santiago de Compostela, Spain }

\address[fic]{CITIC, Computer Architecture Group, Universidade da Coru\~na, 15071 A Coru\~na, Spain}

\begin{abstract}
Today we live in a context in which devices are increasingly interconnected and sensorized and are almost ubiquitous. Deep learning has become in recent years a popular way to extract knowledge from the huge amount of data that these devices are able to collect. Nevertheless, centralized state-of-the-art learning methods have a number of drawbacks when facing real distributed problems, in which the available information is usually private, partial, biased and evolving over time. Federated learning is a popular framework that allows multiple distributed devices to train models remotely, collaboratively, and preserving data privacy. However, the current proposals in federated learning focus on deep architectures that in many cases are not feasible to implement in non-dedicated devices such as smartphones. Also, little research has been done regarding the scenario where data distribution changes over time in unforeseen ways, causing what is known as concept drift. Therefore, in this work we want to present \ourMethod{} (\ourAcronym{}), a new federated and continual architecture that uses light, traditional learners. Our method allows powerless devices (such as smartphones or robots) to learn in real time, locally, continuously, autonomously and from users, but also improving models globally, in the cloud, combining what is learned locally, in the devices. In order to test our proposal, we have applied it in a heterogeneous community of smartphone users to solve the problem of walking recognition. The results show the advantages that \ourAcronym{} provides with respect to other state-of-the-art methods.

\end{abstract}

\begin{keyword}
 federated learning \sep continual learning \sep distributed learning \sep semi-supervised classification \sep cloud-based ensemble \sep smartphones.
\end{keyword}

\end{frontmatter}


\section{Introduction}
\label{sec:introduction}
Smartphones, tablets, wearables, robots and ``things'' from the Internet of Things (IoT) are already counted in millions and allow a growing and sophisticated number of applications related to absolutely all human domains: education, health, leisure, travel, banking, sport, social interaction, etc. If we also take into account factors such as the progressive sensorization and the connection to the networks of these devices, we can talk about an authentic ``society of devices'' that is being formed around people. 

The volume of data generated by these agents is growing rapidly. Having such an exponentially growing amount of data collected on real working conditions from distributed environments, together with a good intercommunication between devices~\cite{li2018}, opens up a new world of opportunities. In particular, it will allow devices to incorporate models that evolve and adapt in order to perform better and better, thus benefiting the consumers. 

In the context of distributed devices (smartphones, robots, etc.), applying traditional cloud-centric machine learning processes involves gathering data from all the devices in the cloud. This data, typically sensor measurements, photos, videos and location information, must be later uploaded and processed centrally in a cloud-based server or data center. Thereafter, the data is used to provide insights or produce effective inference models. With this approach, \emph{deep learning} techniques have proven to be very effective in terms of model accuracy~\cite{qiu2016}. However, in this kind of scenarios, cloud-centric learning is usually either ineffective or infeasible, since it involves facing the following challenges: 

\paragraph{Challenge 1: Scalability}
    There are limitations both in storage and communication costs, as well as in computing speeds.
    Central storage and the transfer of huge amounts of data over the network might take extremely much time and also require an unbearable financial cost. Note also that communication may be a continuous overhead, as data form real environments is continuously been updated. This can be especially challenging in tasks involving unstructured data, e.g., in video analytics~\cite{ananthanarayanan2017real}. A cloud-centric approach can also imply long propagation delays and incur unacceptable latency for applications in which real-time decisions have to be made, e.g., in autonomous driving systems~\cite{ananthanarayanan2017real,montanaro2019towards}. Similarly, central computing can take much more time than parallel processing of smaller parts of data.
    \paragraph{Challenge 2: Data privacy and sensitivity}
    Many popular applications deal with sensitive data, that is, any information about an user that circulates on the Internet and allows him/her to be identified. For example, the ID card, telephone number, or address, but also pictures, videos, browsing history, or geolocation. The central collection of such data is not desirable as it puts people's privacy into risk. In recent years, governments have been implementing data privacy legislations in order to protect the consumer. Examples of this are the European Commission's General Data Protection Regulation (GDPR)~\cite{custers2019eu} or the Consumer Privacy Bill of Rights in the US~\cite{gaff2014privacy}. In particular, the consent (GDPR Art. 6) and data minimisation principles (GDPR Art. 5) limit data collection and storage only to what is consented by the consumer and absolutely necessary for processing. \\


These limitations have led in recent years to the emergence of new learning paradigms that bring processing and analysis capacity closer to the devices themselves, following the philosophy of edge or fog computing~\cite{dolui2017,luan2015}. A good example of this is the recent \emph{federated learning} (FL) ~\cite{li2019federated,mcmahan2016federated}, which allows the training of the model across multiple decentralized edge devices holding local data samples, without exchanging them. Nevertheless, there is still plenty of room for improvement in this area. 
For example, most work on FL is based on deep neural network architectures, but deep learning usually requires huge amounts of data and high computational capabilities (not available on the edge devices). Also, there is a number of practical concerns that arise when running federated learning in production. In particular, the problem of nonstationarity and concept drift (i.e., when the underlying data distribution changes over time) is perhaps one of the most important ones, and has not yet been addressed in FL literature. 

Therefore, in the present work, we propose a new FL algorithm, \emph{\ourMethod{}} (\ourAcronym{}). Our method allows powerless devices (such as smartphones, wearables, or robots) to learn in real time, locally, continuously, autonomously and from users, but also improving models globally, in the cloud, combining what is learned locally, in the devices. Basically, it consists on the training of light, weak, local learners, on the devices themselves, which are later consensuated globally, in the cloud, using ensemble techniques. This global model is then returned to the devices so that they speed up the local learning and help to quickly improve device behaviour, at the same time that they are subdued to a new local adaptation process. Our algorithm is also able to deal with continuous single-task problems where the underlying distribution of data might be non-IID (independent and identically distributed) among the different client devices, but that also can change in unforeseen ways over time (concept drift).
As we will describe later, there are important benefits as a consequence of the explicit detection of concept drifts. In short, we can say that it helps us to answer two key questions: \emph{what} needs to be learned and \emph{when} it should be learned. 
In this work, we will show the performance of our approach when it is used to solve the specific semi-supervised classification task of walking recognition in smartphones. 

The rest of the paper is structured as follows: Section~\ref{sec:state-of-the-art} provides a review of the state of the art. In Section~\ref{sec:method-main}, \ourAcronym{} is presented. Section~\ref{sec:local-learning} explains in detail the learning carried out on the devices. Section~\ref{sec:global-learning} exposes the details of the consensus performed in the cloud. Section~\ref{sec:experimental-results} presents the experimental results in walking recognition. Finally, some conclusions are presented in Section~\ref{sec:conclusions}.

\section{Related work}
\label{sec:state-of-the-art}

As we have already mentioned in the previous section, applying cloud-centric learning in networks of smart and distributed agents, such as smartphones, robots or IoT devices, involves a number of issues. We are referring to the two challenges (scalability and data privacy) mentioned above. In this section, we are going to review some of the strategies that are being applied in this context of devices in order to address these challenges: cloud-centric variations, distributed learning, and federated learning. 

\subsection{Cloud-centric variations}
\label{sec:deep-learning}
In the last decade, there have emerged some cloud-centric variations that suggest that the learning process can take place in the cloud, but the learned model can be then transferred to the devices, which is where it is executed~\cite{nakkiran2015,vasilyev2015}. Other proposals suggest that there may even be a final adjustment of the model at the local level. The latter would fall within the transfer learning paradigm~\cite{zhang2019}, which focuses on storing knowledge gained while solving one problem and applying it to a different but related problem. In this case, a global deep learning model obtained in the cloud would be a general solution that could be then adapted to each device locally.
Other hybrid solutions using local and cloud computing have also been explored. For instance, in~\cite{lane2015}, the pre-training of a deep neural network (DNN) is carried out in the smartphone and the subsequent supervised training is performed in the cloud.

However, any of these strategies will still involve moving a significant volume of potentially sensitive data. Therefore, a better option for learning in this kind of scenarios where data is naturally distributed seems to be a decentralized approach. 

\subsection{Distributed learning}
\label{sec:distributed-learning}

\emph{Distributed learning}~\cite{peteiro2013,verbraeken2019survey,gu2019distributed} is not something new. Different from the cloud-centric approach, in this kind of algorithms the learning process can be carried out in a distributed and parallel manner along a network of interconnected devices, a.k.a. \emph{clients} or \emph{agents}. This client devices are able to collect, store, and process their own data locally. The learning process can be performed with or without explicit knowledge of the other agents. Once each device performs its local learning, a global integration stage is typically carried out in the cloud, so that a global model is agreed upon. Allocating the learning process among the network of devices is a natural way of scaling up learning algorithms in terms of storage, communication and computational cost. 
Furthermore, it makes it easier to protect the privacy of the users, since sharing raw data with the cloud or with other participants can be avoided.

Distributed machine learning algorithms can be roughly classified into two different groups: (1) distributed optimization algorithms, and (2) ensemble methods.

\subsubsection{Distributed optimization algorithms}
These methods mainly focus on how to train a global model more efficiently, by managing large scale of devices simultaneously and making proper use of their hardware in a distributed and parallel way. For that purpose, there already exist many powerful optimization algorithms to solve the local sub-problems, which are developed from Gradient Descent~(GD). The best known one is the ancient \emph{Stochastic Gradient Descent}~(SGD), which greatly reduces the computation cost in each iteration compared with the normal gradient descent~\cite{robbins1951stochastic, recht2011hogwild}. 
There are other proposals that rely on other optimization methods, such as those that use augmented Lagrangian methods, being \emph{Alternating Direction Method of Multipliers} (ADMM) the most popular one~\cite{andersson2014,lin2013}. Another example are Newton-like techniques, such as \emph{Distributed Approximate NEwton} (DANE)~\cite{shamir2014communication} or \emph{Distributed Self-Concordant Optimization} (DiSCO)~\cite{zhang2015disco}. 

The main drawback of this kind of algorithms is that they require the devices to share a common representation of the model and feature set. Moreover, most of the proposals that fall into this category are not usually intended to be applied in the context of smartphones and IoT devices, as they typically involve high computing costs in each training step. Thus, they are not suitable for working on this kind of devices as they might have a negative effect on experience of user's daily usage (high battery consumption, overheating, etc.).

\subsubsection{Ensemble methods}
Ensemble learning~\cite{sagi2018} improves the predictive performance of a single model by training multiple models and combining their predictions (e.g., using a voting system) or even the classifiers themselves (e.g., generating a definitive model from different local sets of rules)~\cite{tsoumakas2002,hall2000,guo1999,hall1998}. Thus, the ensemble approach is almost directly applicable to a distributed environment since a local classifier can be trained at each device and then the classifiers can be eventually aggregated using some ensemble strategy. Some examples of the existing proposals that follow this approach are \emph{Effective Stacking}~\cite{tsoumakas2002}, \emph{Knowledge Probing}~\cite{guo1999} or \emph{Distributed Boosting}~\cite{lazarevic2002boosting}.

\subsection{Federated learning}
\label{sec:federated-learning}
In recent years, a new learning framework called \emph{federated learning}~(FL) has been boosted by Google~\cite{konevcny2015federated,mcmahan2016federated,lim2019federated,li2019federated}. Its core idea is very similar to distributed learning, i.e., solving local problems on the devices and aggregate updates on a server without uploading the original user data. In fact, we could even contemplate state-of-the-art federated methods as distributed optimization algorithms (the first group of distributed methods mentioned above). However, FL has been considered in the literature as an independent paradigm as it has some key differences with more traditional distributed learning. Firstly, the main objective of federated learning is to ensure the privacy of the data and the users, so sharing sensitive information is not an option in FL. Besides, both approaches make different assumptions on the properties of the local datasets, as distributed learning mainly focus on parallelizing computing power, where federated learning, essentially designed to work on mobile device networks, aims at training on heterogeneous datasets~\cite{konevcny2015federated}. A common underlying assumption in distributed learning is that the local datasets are identically distributed and roughly have the same size. None of these hypotheses are made for federated learning. Instead, the datasets are typically heterogeneous and their sizes may span several orders of magnitude.

Federated learning seems, at present, the most suitable approach for this context of \emph{multi-device learning}. Nevertheless, there are still some constraints in the state-of-the-art methods that need to be addressed. In this work, we would like to highlight two of them: (1) the non-viability of deep architectures, and (2) the lack of methods to deal with nonstationarity of data. By following the list of challenges initiated in Section~\ref{sec:introduction}, we have:
    
\paragraph{Challenge 3: Deployment of deep federated learning} As far as we are aware, all the existing FL algorithms are based on deep neural network architectures. This is a significant constraint, since the type of learner to be used is often conditioned by the task to be solved, and deep neural networks are not always the best solution. In addition, the use of DNNs usually demands high amounts of data and high computing capabilities. To date, client devices, such as smartphones, do not have the appropriate hardware and software to carry out this computing. The most popular deep learning frameworks and libraries, such as TensorFlow~\cite{abadi2016tensorflow} or Pytorch~\cite{paszke2019pytorch}, still do not have any available version for operating systems such as Android or iOS that allow the training of these models. In terms of hardware, although much progress has been made in recent decades, these kind of devices still cannot compete with dedicated computers capable of getting the most out of their GPUs. This is a major impediment to the introduction of FL-based solutions in the real world. Nevertheless, in many cases it could be solved using traditional learning methods, such as decision trees or support vector machines, that have lightweight implementations in multiple programming languages and for multiple operating systems.
   
\paragraph{Challenge 4: Nonstationarity}
Most machine learning algorithms assume that data is stationary and IID, i.e., independent and identically distributed. However, it does not hold in many real world applications, where the underlying distribution of the data streams changes over time and between the different devices. Some authors have evaluated some FL algorithms on  non-IID scenarios, analyzing the impact of non-identical client data partitions~\cite{mcmahan2016federated, zhao2018federated, caldas2018leaf}. Nevertheless, little research has been done studying the effect of non-stationary data streams along time. This phenomenon is known in the literature as \emph{concept drift}~\cite{widmer1996}. If the concept drift occurs, the inducted pattern of past data may not be relevant to the new data, leading to poor predictions or decision outcomes. 
Yoon et al.~\cite{yoon2020federated} propose a method for federated continual learning with Adaptive Parameter Communication, Fed-APC, which additively decomposes the network weights into global shared parameters and sparse task-specific parameters. They validate their approach on several settings, including multi-incremental-task continual scenarios.
However, to the best of our knowledge, there are no other researchers working on federated learning who have addressed concept drift detection and adaptation in the underlying distribution of data. 

The new federated method we propose in this paper, \ourAcronym{}, is intended to overcome these challenges. On the one hand, it is designed to work with weak traditional learners, e.g. decision trees, instead of DNNs. On the other hand, our algorithm is also able to deal with concept drift in continuous single-task problems. 
Moreover, the proposed architecture is asynchronous, and contemplates a semi-supervised learning, two additional aspects that make it more suitable to be deployed in real-world scenarios.

\section{\ourMethod{}: an overview}
\label{sec:method-main}

\begin{figure}[htbp]
	\centering
	\includegraphics[width=1.0\linewidth]{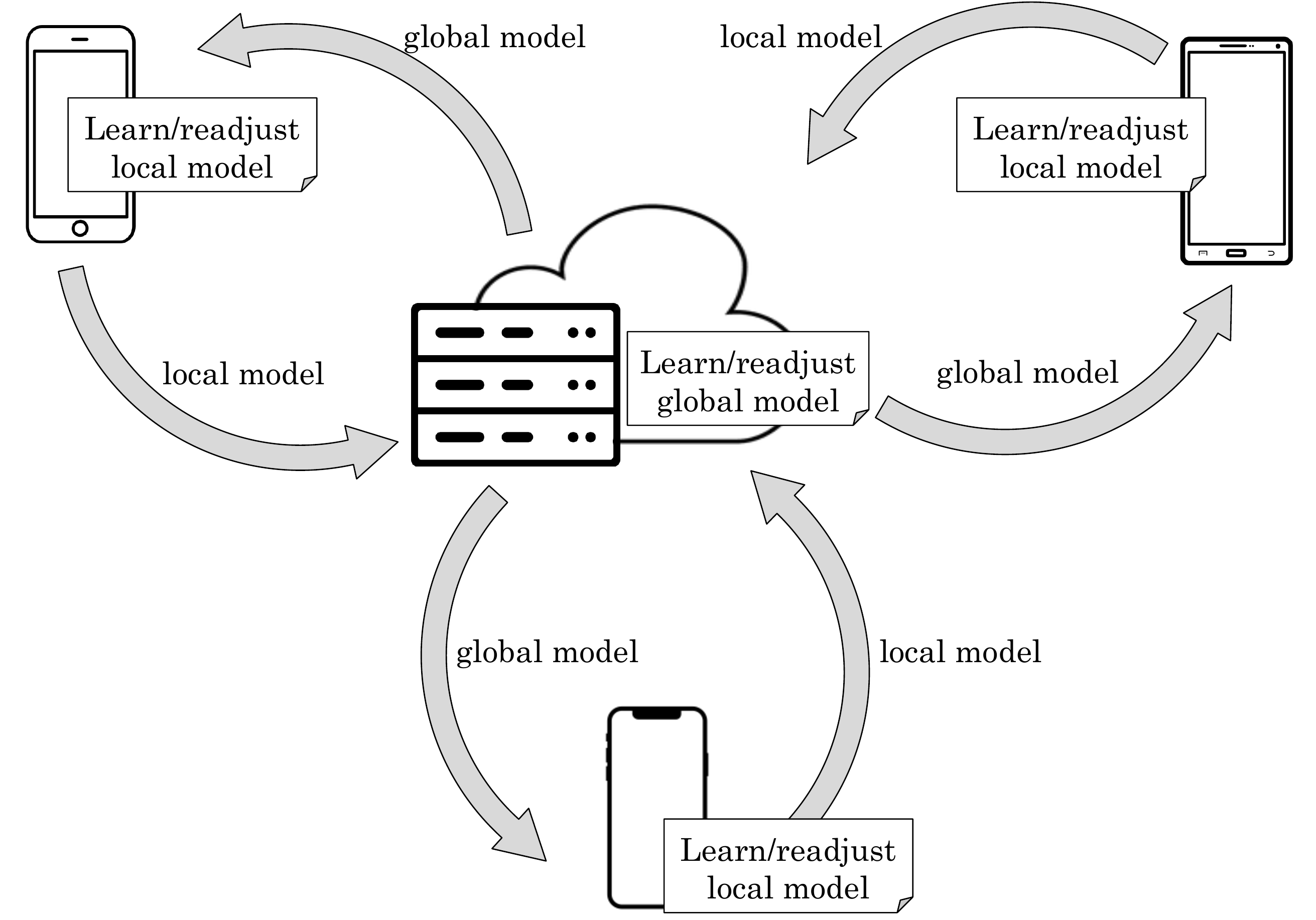}
	\caption{Diagram of the proposed architecture.} 
	\label{fig:method-architecture}
\end{figure}

Figure~\ref{fig:method-architecture} shows a high-level diagram of our proposal. As we can see, \ourAcronym{} has a cyclical architecture. In the figure there are smartphones, but we could think of any other set of agents, either homogeneous or heterogeneous, including tablets, wearables, robots, or IoT devices. Each client device is able to perceive its environment through its sensors and is connected to the cloud.

Iteratively, each of the devices creates and refines its own local model of the learning problem that is intended to be solved. For that, devices are continuously acquiring and storing new information through their sensors. This information is raw data, which must be locally preprocessed before being able to use it in a learning stage: noise detection, data transformation, feature extraction, data normalization, instance selection, etc. When local models are obtained, they are sent to the cloud where a new learning stage is performed to join the local knowledge, thus obtaining a global model. The global model is then shared with all the devices in the network. Each device can take advantage of that global model to improve the local one, thus starting a new cycle.
New data is continuously recorded in each device and will be also used to retrain better local models or refine existing ones. Of course, an improvement of the local models will result in an improvement at the global level too. Note that the information available at the local level will increase progressively. As it is unrealistic to assume that infinite storage is available, a compromise solution must be reached that retains previously learned knowledge that is still relevant at the same time as old information is replaced with new knowledge, avoiding the catastrophic forgetting.

In this work we have focused on semi-supervised scenarios, where the data in the local devices is partially labeled.
Nevertheless, the application of \ourAcronym{} to fully supervised problems is immediate. In the following sections we explain the details of our method at both learning levels, local (Section~\ref{sec:local-learning}), and global (Section~\ref{sec:global-learning}). However, it should be borne in mind that this kind of architecture would allow the integration of other methods different from those we propose below. Indeed, as we will see later, we believe that there is room for improvement at both levels.

\section{Local learning}
\label{sec:local-learning}

\begin{figure}[htb]
	\centering
	\includegraphics[width=0.94\linewidth]{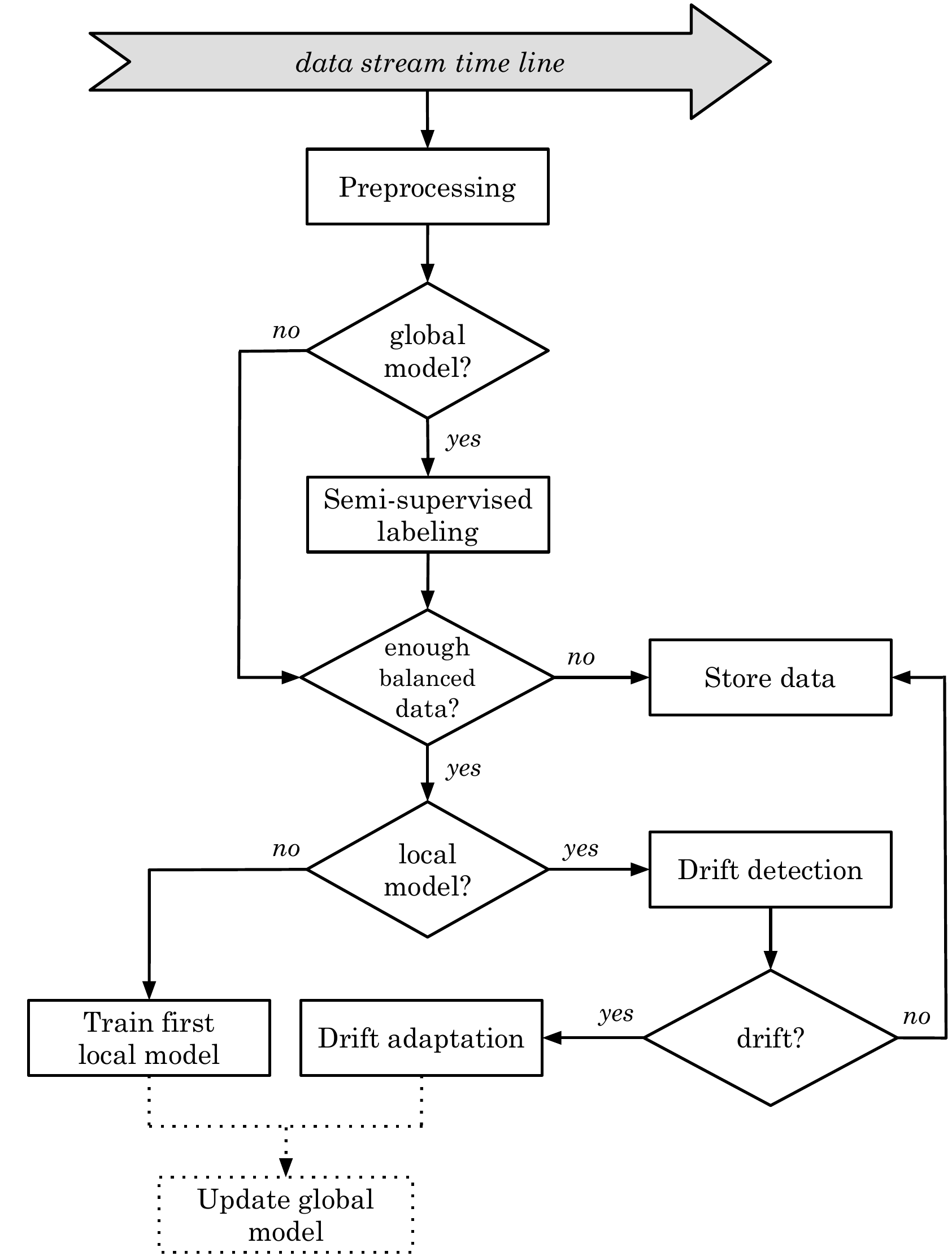}
	\caption{Work flow on a local device.} 
	\label{fig:local-workflow}
\end{figure}

Figure~\ref{fig:local-workflow} shows the continuous work flow for each device. Basically, devices gather data from the environment. This data, conveniently preprocessed, is used to build or update a local model. The preprocessing of the data refers to feature extraction, normalization, instance selection, etc. As we deal with a semi-supervised task, this data will be partially annotated. 

In order to learn the local model, in this work we have opted for the use of ensemble methods. These methods have received much attention in stream data mining and large-scale learning. In particular, every device builds an ensemble of base classifiers. 
Any algorithm that provides posterior probabilities for its predictions can be used as base classifier. In our case, in our experimental results (Section~\ref{sec:experimental-results}), we tried different methods as base classifiers: Na\"ive Bayes, Generalized Linear Models (GLM), C5.0 Decision Trees,  Support Vector Machines (SVM), Random Forests and  Stochastic Gradient Boosting (SGB).
In the same way, any state-of-the-art method could be used to combine the predictions of the base classifiers of the local ensemble. We opted for a simple but effective approach, employing decision rules, which effectively combine a posteriori class probabilities given by each classifier. Rule based ensemble have received very much attention because of their simplicity and because they do not require training~\cite{czyz2004, kittler1998, tumer1996}. When the base classifiers operate in the same measure space, as it is this case, averaging the different posterior estimates of each base classifier reduces the estimation noise, and therefore improves the decision~\cite{tumer1996}. Thus, we should use a rule that averages the posterior estimates of the base classifiers. The sum rule could be used, but if any of the classifiers outputs an a posteriori probability for some class
which is an outlier, it will affect the average and this could lead to an incorrect decision. It is well known that a robust estimate of the mean is the median. Thus, we use the \emph{median rule}. The median rule predicts that an instance $x$ belongs to class $c_j$ if the following condition is fulfilled:
\begin{equation}
\label{eq:median-rule}
    \resizebox{0.91\linewidth}{!}{%
     $\textnormal{median} \{y_{1_j}(x), \dots, y_{N_j}(x)\} = \max_{k=1}^{C} \textnormal{median} \{y_{1_k}(x), \dots, y_{N_k}(x)\},$%
     }
\end{equation}
where $C$ is the number of possible classes ($c_1,c_2, \dots, c_C$), $N$ is the number of base classifiers and $y_i = \{y_{i_1}(x), \dots, y_{i_C}(x)\}$ is the output of the $i$-th classifier, for $i = 1, \dots, N$.

As we can see in Figure~\ref{fig:local-workflow}, our local device gathers data and, if there is a global model available, it uses it to annotate the unlabeled data (semi-supervised transduction). Otherwise, it keeps collecting data until it has enough examples to perform the training of the first local model.
A common question in machine learning is how much data is necessary to train a model. Unfortunately, the answer is not simple. In fact, it will depend on many factors, such as the complexity of the problem and the complexity of the learning algorithm.
Significant research efforts have been made to find the relationship between classification error, learning sample size, dimensionality and complexity of the classification algorithm~\cite{jain1982,kanal1971,raudys1991}. Statistical heuristic methods have been frequently used to calculate a suitable sample size, typically based on the number of classes, the number of input features or the number of model parameters.
To ensure a minimally balanced training set with representation of all classes, in this work we establish a single heuristic rule that must be fulfilled in order to allow the training of a local model. We define a minimum amount of labeled data, $L$, so that there must be at least $\frac{L}{2C}$ examples from each class in the training set to allow the training process, where $C$ is the number of possible classes. 
The first base model of the local ensemble --Equation~\eqref{eq:median-rule}-- can be trained as soon as this rule is met.  
In our experiments we have used $L = 2\Delta$, where $\Delta$ is a parameter used in the drift detection algorithm that we will expose in Section~\ref{sec:drift-detection}. We define $L$ in terms of $\Delta$ as both parameters are strongly related.
%

As we pointed out before, the local device collects data which is partially annotated. In fact, it will be common to have a much greater number of unlabeled than labeled examples. In order to deal with this, we perform what is called \emph{semi-supervised transduction}. To understand how it works we must remind that in  \ourAcronym{} all devices receive the last global model achieved by consensus in the cloud (Figure~\ref{fig:method-architecture}). 
To take advantage of that knowledge globally agreed, our proposal uses the global model for labeling data that has no label. As soon as a new unlabeled instance is available, we use the latest global model, if any, to predict a possible label and, then, we filter the predictions based on their degree of confidence. We define the confidence of a prediction as the classifier conditional posterior probability~\cite{duin1998}, i.e., the probability $\textnormal{P}(c_i | x)$ of a class $c_i$ from one of the $C$ possible classes $c_1, c_2, \dots, c_C$ to be the correct class for an instance $x$. 
It  is a normalized value between 0 and 1. 
We accept the predicted label as the real label of the example when its confidence is equal to or greater than a threshold $\gamma$, whose optimal value we have empirically set at $\gamma = 0.9$. Low thresholds ($\gamma < 0.8$) may introduce a lot of noise in the training set, while very high thresholds ($\gamma >= 0.95$) may allow to add very few examples to the labeled set. Therefore, we consider that $\gamma = 0.9$ is an adequate value.

Finally, once the device has a global model and a local model, the remaining question is when this local model should be updated using the new data that is being collected. It makes no sense to update the local model if it is performing well. However, as we said before, data is usually non-stationary, i.e., its distribution evolves in time in an unpredictable way. This is what is usually called \textit{concept drift}. If a concept drift happens the model will lower its performance. Thus, as we will show in Section~\ref{sec:drift}, we will update the local model when a concept drift is detected.

\subsection{Learning under concept drift}
\label{sec:drift}

Concept drift is a continual learning issue~\cite{lesort2020continual}. Formally, we can define it as follows: Given a time period $[0,t]$, a set of samples, denoted as $S_{0,t} = \{s_0, \dots, s_t\}$, where $s_i = (x_i, y_i)$ is one observation (or data instance), $x_i$ is the feature vector, $y_i$ is the label, and $S_{0,t}$ has a certain joint probability distribution of $x$ and $y$, $P_{t}(x,y)$. Therefore, concept drift can be defined as a change in the joint probability at timestamp $t$, such that  $\exists t: P_{t}(x,y) \neq P_{t+1}(x,y)$.

The aforementioned is the definition of conventional concept drift. Nevertheless, in a federated learning setting, with multiple clients and a global server, we must adapt it a little. Now, the goal is to train a global model in a distributed and parallel way using the local data of the $D$ available clients. Each client will have a different bias because of the conditions of its local environment, and likewise, its data stream may change in different ways over time. Therefore, there may occur concept drifts that affect all clients, some of them, or just a single one. Thus, we can generalize the problem in the following way: Given a time period $[0,t]$, a set of clients $\{d_1, \dots,d_D\}$ , and a set of local samples for each client, denoted as $S_{0,t}^k = \{s_0^k, \dots, s_t^k\}$, where $s_i^k = (x_i^k, y_i^k)$ is one data instance from client $d_k$, $x_i^k$ is the feature vector, $y_i^k$ is the label. 
Each local dataset $S_{0,t}^k$ has a certain joint probability distribution $P_{t}^k(x,y)$. 
A \emph{local concept drift} occurs at timestamp $t$ for client $d_k$ if  $\exists t,c: P_{t}^k(x,y) \neq P_{t+1}^k(x,y)$.

However, note that a local concept drift does not necessarily have a direct impact on the global federated model. It may be the case that a local drift on device $d_k$ will result in a change in the distribution of $d_k$, but not in the joint distribution of all clients, $P_{t}^G(x,y)$. In that case, the federated model will not be affected by the local change and this one can be ignored. Thus, we can define a \emph{global concept drift} as a distribution change at time $t$ in one or more clients $\{d_k, \dots,d_l\} \subseteq \{d_1, \dots,d_C\}$ such that $\exists t: P_{t}^G(x,y) \neq P_{t+1}^G(x,y)$.
If a global concept drift happens the model might lower its performance, so we need to detect these global changes and adapt to them.

Research on learning under concept drift is typically classified into three components~\cite{lu2018}: (1) drift detection (whether or not drift occurs), (2) drift understanding (when, how, where it occurs) and (3) drift adaptation (reaction to the existence of drift). In this work, we have focused on drift detection and adaptation. 

\subsubsection{Drift detection}
\label{sec:drift-detection}
Drift detection refers to the techniques and mechanisms that characterize and quantify concept drift via identifying change points or change intervals in the underlying data distribution. In our proposal, if a concept drift is identified it means that the local model is no longer a good abstraction of the knowledge of the device, so it must be updated.
Drift detection algorithms are typically classified into three categories~\cite{lu2018}: (1) error rate-based methods, (2) data distribution-based methods and (3) multiple hypothesis test methods. The first class of algorithms focuses on tracking changes in the online error rate of a reference classifier (in our case, it would be the latest local model). The second type uses a distance function or metric to quantify the dissimilarity between the distribution of historical data and the new data. The third group basically combines techniques from the two previous categories in different ways. Error rate-based methods operate only on true labeled data, because they need the labels to estimate the error rate of the classifier. Therefore, in order to take advantage of both labeled and unlabeled data, in this work we decided to use a data distribution-based algorithm, which do not present such a restriction. Thus, we have used an adapted version of the change detection technique (CDT) originally proposed in~\cite{haque2016}, which is a CUSUM-type CDT on \textit{beta} distribution~\cite{baron1999convergence}. Algorithm~\ref{alg:drift} outlines the proposed method.

\SetKw{kwInput}{Input:}
\SetKw{kwOutput}{Output:}
\begin{algorithm}[!htb]
            \kwInput{$W$, $N$, $x$, $T_h$, $\alpha$, $\Delta$, $N_{max}$} \\
            \kwOutput{$W$, $N$} \\
			    $[\hat{y}, \varsigma_x] \leftarrow \textnormal{classify}(x)$ \\
			    $w_{new} \leftarrow [x,\varsigma_x]$ \\
			    $W \leftarrow W \cup w_{new}$ {\footnotesize// Add the pattern and its confidence into $W$} \\
			    $N \leftarrow N+1$ \\
			    \If{$N >= N_{max}$}{
    				    $w_0 \leftarrow \varnothing$  {\footnotesize// Remove the oldest element in $W$} \\
    				    $N \leftarrow N-1$ \\
    			}
			    $s_f \leftarrow 0$ \\
			    $r \leftarrow \textnormal{random}(0,1)$ {\footnotesize// Generate random number in the interval [0,1]} \\
			    \If{$e^{-2\varsigma_x} \geq r$}{ 
    				\For{$k \leftarrow \Delta$ \KwTo $N - \Delta$}{
    				    $m_b \leftarrow \textnormal{mean}(\varsigma_1  : \ \varsigma_k \ \in W)$ \\
    				    $m_a \leftarrow \textnormal{mean}(\varsigma_{k+1}  : \ \varsigma_N \ \in W)$ \\
    				    \If{$m_a \leq (1- \alpha) \cdot m_b$}{
    				        $s_k \leftarrow 0$ \\
    				        $[\hat{\alpha}_b, \hat{\beta}_b] \leftarrow \textnormal{estimateParams}(\varsigma_1  : \ \varsigma_k \ \in W)$  \\
    				        $[\hat{\alpha}_a, \hat{\beta}_a] \leftarrow \textnormal{estimateParams}(\varsigma_{k+1}  : \ \varsigma_N \ \in W.)$ \\
        					\For{$i \leftarrow k+1$ \KwTo $N$}{
        					    $s_k \leftarrow s_k + \log\left(\frac{f\left(\varsigma_i \in w_i \ | \ \hat{\alpha}_a, \ \hat{\beta}_a\right)}{f\left(\varsigma_i  \in w_i  \ | \ \hat{\alpha}_b, \ \hat{\beta}_b\right)}\right)$ \\
        					}
        					\If{$s_k > s_f$}{
        						$s_f \leftarrow s_k$ \\
    						}
    					}
    				}
    				\If{$s_f > T_h$}{
    				    \textnormal{driftAdaptation}($W$) {\footnotesize// See details in Section~\ref{sec:drift-adaptation}} \\
    				    $W \leftarrow \{\varnothing\}$ {\footnotesize// Reinitialize the sliding
    				    window} \\
    				    $N \leftarrow 0$
    				}
				}
	\caption{Change detection algorithm.\label{alg:drift}}
\end{algorithm}

As soon as a new instance $x$ is available, Algorithm~\ref{alg:drift} is invoked. First of all, the confidence of the current local classifier on that instance, $\varsigma_x$, is estimated (line 3 in the pseudocode) and both, instance and confidence, are stored in a sliding window $W$ of length $N$ (lines 4 to 6). Note that $W = \{w_1, w_2, \dots, w_N\}$ is a history of tuples of the form ``[instance,~confidence]'', so that $W = [X,\varSigma]$, where $X = \{x_1, x_2, \dots, x_N\}$ is the history of instances and $\varSigma = \{\varsigma_1, \varsigma_2, \dots, \varsigma_N\}$ the associated confidence vector. 
We will try to detect changes in the confidence of the predictions provided by the current local model for the patterns in $W$. In the original method proposed in~\cite{haque2016}, authors do not use a sliding window, but a dynamic window, which is reinitialized every time a drift is identified. However, they do not establish any limits on the size of this dynamic window, which is not very realistic, since its size could grow to infinity if no drift is detected. Therefore, in our proposal, we use a sliding window and we set a maximum  size $N_{max} = 20\Delta$, where $\Delta$ is a strongly related parameter used in the core of the CDT algorithm, as we will explain below. Once this maximum size is reach, adding a new element to $W$ implies deleting the oldest one (lines 7 to 9).

The core of this CDT algorithm (lines 14 to 32) can be a bottleneck in our system if we have to execute it after inserting each confidence value in $W$. Therefore, as shown in Figure~\ref{fig:local-workflow}, CDT will be invoked only if $W$ contains a representative number of instances $X \in W$, i.e., there are at least $\frac{L}{2C}$ labeled instances for each class. However, once this condition is met, CDT would be invoked for every new recorded sample. Therefore, we restrict the number of executions, so that the core of Algorithm~\ref{alg:drift} will be executed with a probability of $e^{-2\varsigma_x}$ (line 13 in the pseudocode). Hence, the higher the confidence, the lower the probability of executing CDT, and \textit{vice versa}.

In the core of the CDT algorithm, $W$ is divided into two sub-windows for every pattern $k$ between $\Delta$ and $N - \Delta$ (lines 14 to 16 in Algorithm~\ref{alg:drift}). 
Let $W_a$ and $W_b$ be the two sub-windows, where $W_a$ contains the most recent instances and their confidences. Each sub-window is required to contain at least $\Delta$ examples to preserve statistical properties of a distribution. When a concept drift occurs, confidence scores are expected to decrease. Therefore, only changes in the negative direction are required to be detected. In other words, if $m_a$ and $m_b$ are the mean values of the confidences in $W_a$ and $W_b$ respectively, a change point is searched only if $m_a \leq (1-\alpha) \times m_b$, where $\alpha$ is the sensitivity to change (line 17). Same as in~\cite{haque2016}, we use $\alpha = 0.05$ and $\Delta = 100$ in our experiments, which are also widely used in the literature.

We can model the confidence values in each sub-window, $W_a$ and $W_b$,  as two different \textit{beta} distributions. However, the actual parameters for each one are unknown. The proposed CDT algorithm estimates these parameters at lines 19 and 20. Next, the sum of the log likelihood ratios $s_k$ is calculated in the inner loop between lines 21 and 23, where $f\left(\varsigma_i \in w_i \ | \ \hat{\alpha},  \hat{\beta}\right)$ is the probability density function (PDF) of the \textit{beta} distribution, having estimated parameters $\left(\hat{\alpha}, \hat{\beta}\right)$, applied on the confidence $\varsigma_i$ of the tuple $w_i = [x_i, \varsigma_i] \in W$. 
This PDF describes the relative likelihood for a random variable, in this case the confidence $\varsigma$, to take on a given value, and it is defined as:
\begin{equation}
\label{eq:pdf-beta}
    f\left(\varsigma  | \ \alpha,  \beta\right) = \begin{cases}
    \frac{\varsigma^{\alpha-1}(1-\varsigma)^{\beta-1}}{B\left(\alpha, \beta\right)}, & \text{if } 0 < \varsigma < 1 \\
    0, & \text{otherwise,} 
  \end{cases} 
\end{equation}
where
\begin{equation}
\label{eq:pdf-beta2}
 B\left(\alpha,  \beta\right) = \int_{0}^{1} \varsigma^{\alpha-1}(1-\varsigma)^{\beta-1} d\varsigma.
\end{equation}

The variable $s_k$ is a dissimilarity score for each iteration $k$ of the outer loop between lines 14 and 28. The larger the difference between the PDFs in $W_a$ and $W_b$, the higher the value of $s_k$~(line 22).
Let $k_{max}$ is the value of $k$ for which the algorithm calculated the maximum $s_k$ value where $\Delta \leq k \leq N - \Delta$. Finally, a change is detected at point $k_{max}$ if $s_{k_{max}} \equiv s_f$ is greater than a prefixed threshold $T_h$~(line 29). As in the original work, we use $T_h = - \log(\alpha)$. 
In case a drift is detected, a drift adaptation strategy is applied~(line 30). We will discuss this strategy in detail in Section~\ref{sec:drift-adaptation}. Moreover, the sliding window $W$ is reinitialized (lines 31 and 32).

\subsubsection{Drift adaptation} 
\label{sec:drift-adaptation}
Once a drift is detected, the local classifier should be updated according to that drift. There exist three main groups of drift adaptation (or reaction) methods~\cite{lu2018}: (1) simple retraining, (2) ensemble retraining and (3) model adjusting. The first strategy is to simply retrain a new model combining in some way the latest data and the historical data to replace the obsolete model. The second one preserve old models in an ensemble and when a new one is trained, it is added to the ensemble, so that the local model itself is the ensemble. The third approach consist of developing a model that adaptatively learns from the changing data by partially updating itself. This last strategy is arguably the most efficient when drift only occurs in local regions. However, online model adjusting is not straightforward and it will depend on the specific learning algorithm being used. In fact, most of the methods from this category are based on decision trees because trees have the ability to examine and adapt to each sub-region separately. 

In our case, we apply ensemble retraining. In particular, as we already mentioned before, we propose a simple rule based ensemble using the median rule --Equation~\eqref{eq:median-rule}-- for local adaptation to concept drift. Each local device is allowed to keep up to $M_l$ local models, which will make up its ensemble. When a drift is detected, a new model will be trained using only the labeled data stored in the dynamic window $W$ described before (Section~\ref{sec:drift-detection}). This new model will be added to the ensemble. If there are already $M_l$ models in the ensemble, the new one replaces the oldest, thus ensuring that there are at most $M_l$ models at any time. Therefore, with this strategy, we also face the infinite length problem, as a constant and limited amount of memory will be enough to keep both training data and the ensemble. In our experiments we used $M_l = 5$.

Therefore, we conceive each local model as an ensemble of base historical models. From now on, for simplicity, we will refer to this ensemble simply as the local model of the device. Note that the estimated posterior probability of the ensemble from Equation~\eqref{eq:median-rule} is the confidence of the local model, that we use for drift detection. 

\section{Global learning}
\label{sec:global-learning}
Every time that a local device train or update its local model, the changes in that local model will be reported to the cloud. For the construction of the global model, we decided once again to use an ensemble, because of the good properties of these kind of methods that we already mention in Section~\ref{sec:local-learning}. As in the local level, any state-of-the-art ensemble technique could be used to combine the predictions of the local models.
Stacking, or stacked generalization~\cite{wolpert1992}, is a technique for achieving the highest generalization accuracy that has probably been the most widely applied in distributed learning~\cite{chan1993,peteiro2013,tsoumakas2002}. Stacking tries to induce which base classifiers (in our case, local models) are reliable and which are not by training a meta-learner. This meta-learner is a higher-level model which is trained on a meta-dataset composed from the outputs of all base classifiers on a given training set.
However, it presents a great limitation, which is the need to have this meta-dataset available in advance. In the context of devices, this would involve gathering a certain amount of labeled data from all users in the cloud before being able to create a global model.
Therefore, a better solution is to use a simpler but equally effective rule-based ensemble, as we already do at the local level (Section~\ref{sec:local-learning}). In this case, the optimal combining rule is the \emph{product rule}, because each local classifier operates in a different measure space~\cite{kittler1998}: different environments, sensors, users, etc.
The product rule predicts that an instance $x$ belongs to class $c_j$ if the following condition is met:
\begin{equation}
\label{eq:product-rule}
    \prod_{i=1}^{N} y_{i_j}(x) = \max_{k=1}^{C} \prod_{i=1}^{N} y_{i_k}(x),
\end{equation}
where $C$ is the number of possible classes ($c_1,c_2, \dots, c_C$), $N$ is the number of base classifiers and $y_i = \{y_{i_1}(x), \dots, y_{i_C}(x)\}$ is the output of the $i$-th classifier, for $i = 1, \dots, N$.
Note that the outputs of this global ensemble are the estimated posterior probabilities that we use on each local device as a confidence measure to decide which unlabeled patterns can and cannot be labeled.

Although the time complexity on the server side is linear, $O(N)$, including all local models in the ensemble is not the best option for several reasons. First, depending on the problem, there could be thousands or millions of connected devices, so using an ensemble of those dimensions could be computationally very expensive. As the global model is sent back to local devices, it would also have a negative impact on the bandwidth and computational requirements of the nodes. In addition, assuming that there is an optimal global abstraction of knowledge, not all the local models will reflect such knowledge in equal measure or bring the same wealth to the ensemble. On the contrary, there will be devices that, accidentally or intentionally, may be creating local models totally separated from reality, which should be detected in time so as not to participate in the ensemble.

Because of all of this, we propose to keep a selection of the  $M_g$ best local models to participate in the global ensemble. In this way, we can know \textit{a priori} the computational and storage resources we will need more easily. When a new local model is received on the cloud, if the device owner of that model is already present in the global ensemble, we can simply update the global ensemble with the new version of the local model. Otherwise, it must compete against the $M_g$ models that are currently in the global ensemble. The server will keep a score representing the relevance of each of the $M_g$ models and will compute that score for each new incoming model. For the computation of the scores we base on the Effective Voting (EV) technique~\cite{tsoumakas2004}. EV propose to perform 10-fold cross validation for the evaluation of each model and then apply a paired t-test for each pair of models to evaluate the statistical significance of their relative performance. Finally, the most significant ones are selected. 
\begin{figure*}[hbtp]
	\centering
	\includegraphics[width=0.54\linewidth]{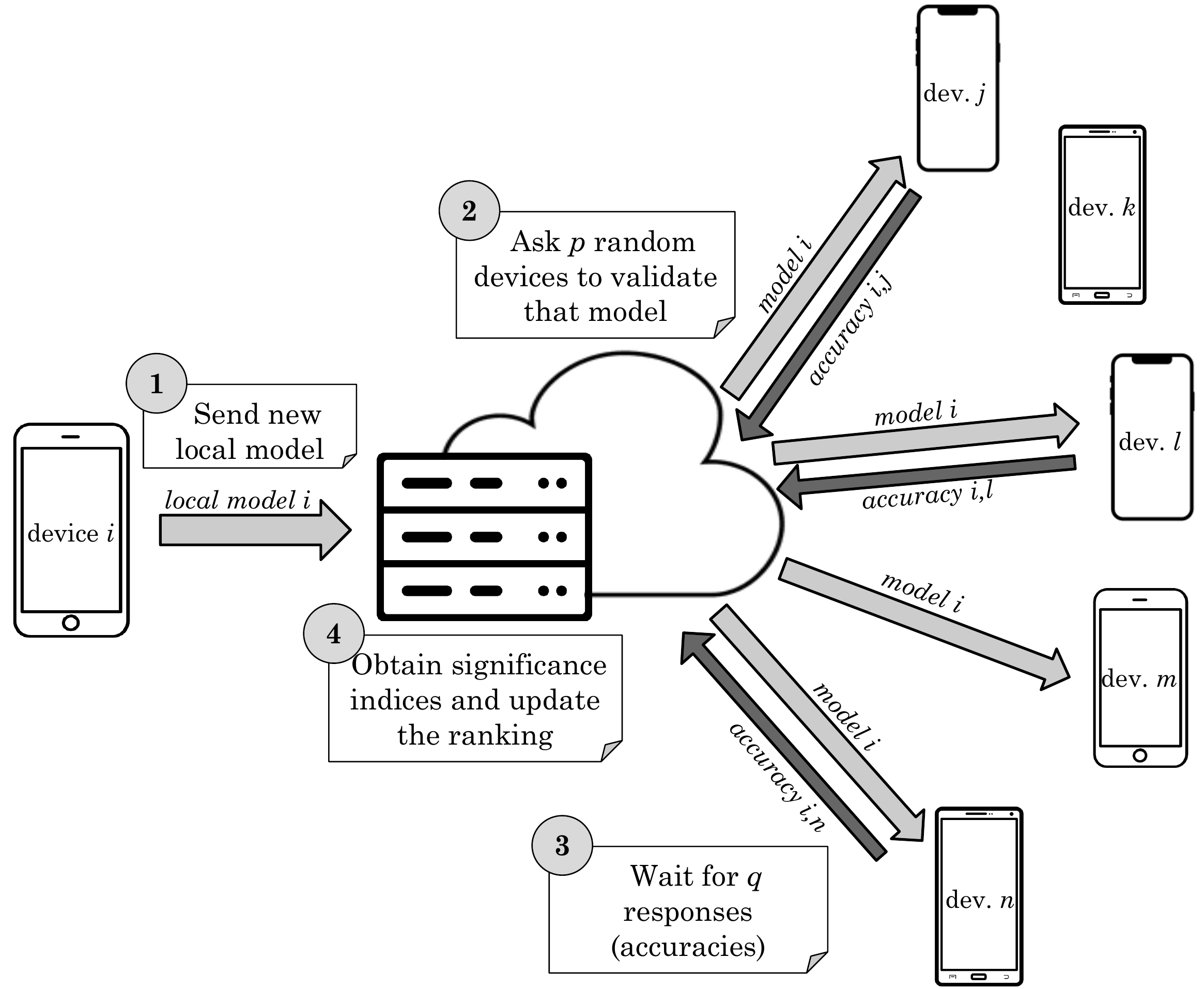}
	\caption{Work flow of the Distributed Effective Voting.} 
	\label{fig:effective-voting}
\end{figure*}

In our context, a cross-validation is not a fair way to evaluate each local model due to the skewness and bias inherent in the distributed paradigm. Thus, when a new local model arrives, 
the server choose $p$ different local devices, randomly selected, and ask them to evaluate that classifier on their respective local training sets. Once this evaluation is done, each device sends back to the cloud its accuracy. Assuming that not all the $p$ selected devices are necessarily available to handle the request, the server waits until it has received $q$ performance measures for that model, being $q \leq p$. This process could be considered a distributed cross-validation. After gathering the $q$ measurements for the current $M_g$ models and the new one, a paired t-test with a significance level of $0.05$ is performed for each pair of models $c_i,c_j$ so that:
\begin{equation}
    \label{eq:t-test}
    t(c_i,c_j) = \begin{cases}
                        1  &\text{if $c_i$ is significantly better than $c_j$}, \\
                       -1  &\text{if $c_j$ is significantly better than $c_i$}, \\
                        0  &\text{otherwise}.
                    \end{cases}
\end{equation}
Then, for each model we calculate its overall significance index:
\begin{equation}
    \label{eq:significance-index}
    S(c_i) = \sum_{j=1}^N t(c_i, c_j).
\end{equation}
Finally, we select the new $M_g$ models with the highest significance index or score, $S$. If there are ties, we break them by selecting the most accurate ones (we compute the mean accuracy from the $q$ available evaluations). 
Figure~\ref{fig:effective-voting} summarizes the whole process, that we can call Distributed Effective Voting (DEV).

In our experiments we just had 10 different devices, so we tried several values for $M_g$, from 3 to 10. However, the size of ensemble usually ranges from 5 to 30 models and it will be strongly dependent on the problem. The $p$ and $q$ sizes depend both on the size of the ensemble ($M_g$) and the total number of devices available online ($N$), so that $M_g \leq q \leq p \leq N$. For simplicity, in our experiments we always used $p = q = M_g$.

\section{Experimental results}
\label{sec:experimental-results}

The aim of this section is to evaluate the performance of \ourAcronym{}. In particular, we are interested in checking to what extent the global models obtained from the consensus of the different local devices are capable of solving distributed and semi-supervised classification tasks. In addition, we also want to evaluate the evolution of these models over the time. In our experiments, we have used several base classifiers at the local level so as not to link the results and the architecture to any particular method (although, as we will see below, some of them are clearly suboptimal according to the strategy we propose).

\begin{figure}[htbp]
\centering
\subfloat[]{\label{fig:steps-example}\includegraphics[width=0.77\linewidth,trim={0.15cm 0.6cm 0.85cm 2.08cm},clip]{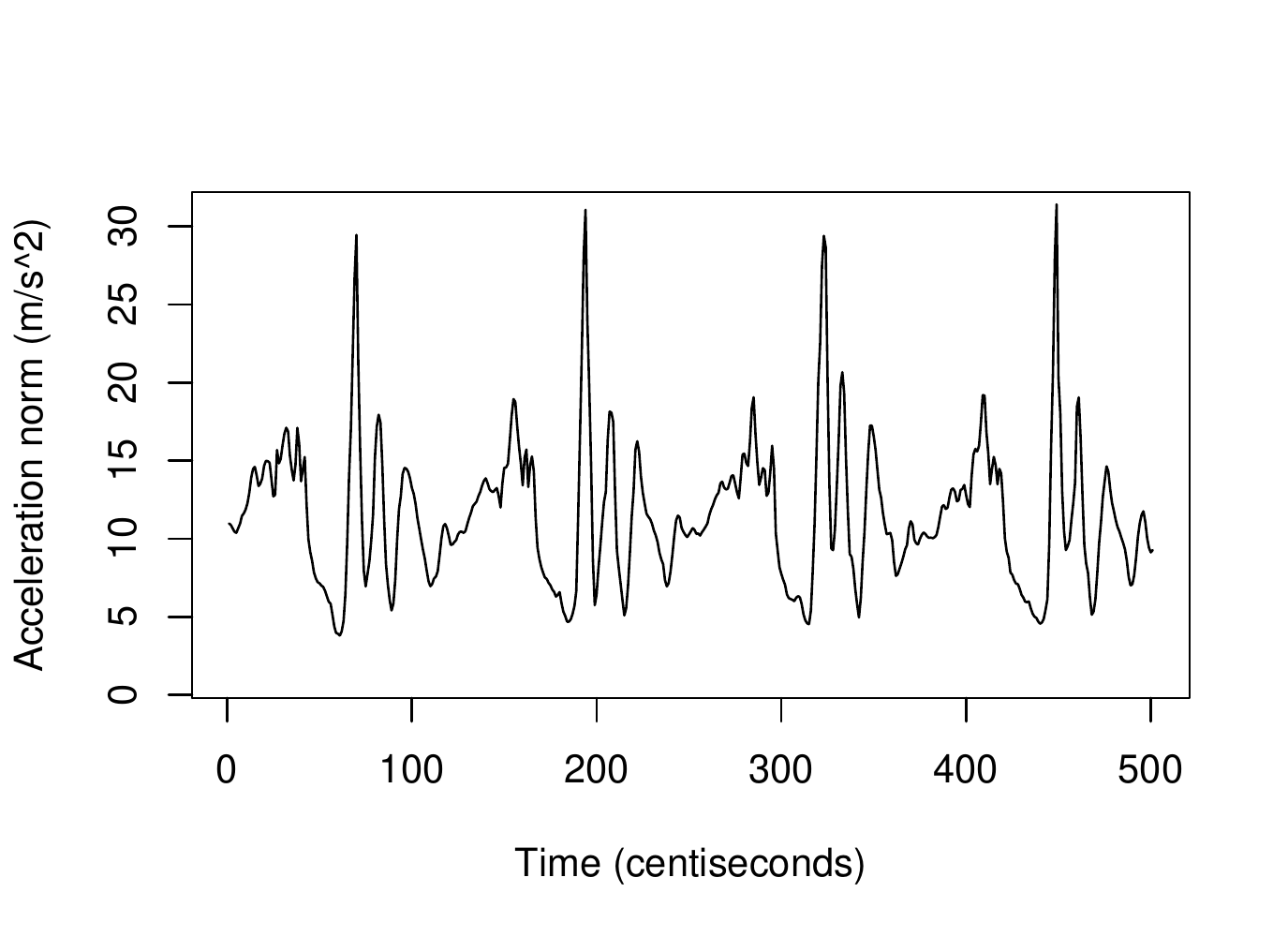}} \\
\subfloat[]{\label{fig:noise-example}\includegraphics[width=0.77\linewidth,trim={0.15cm 0.6cm 0.85cm 2.08cm},clip]{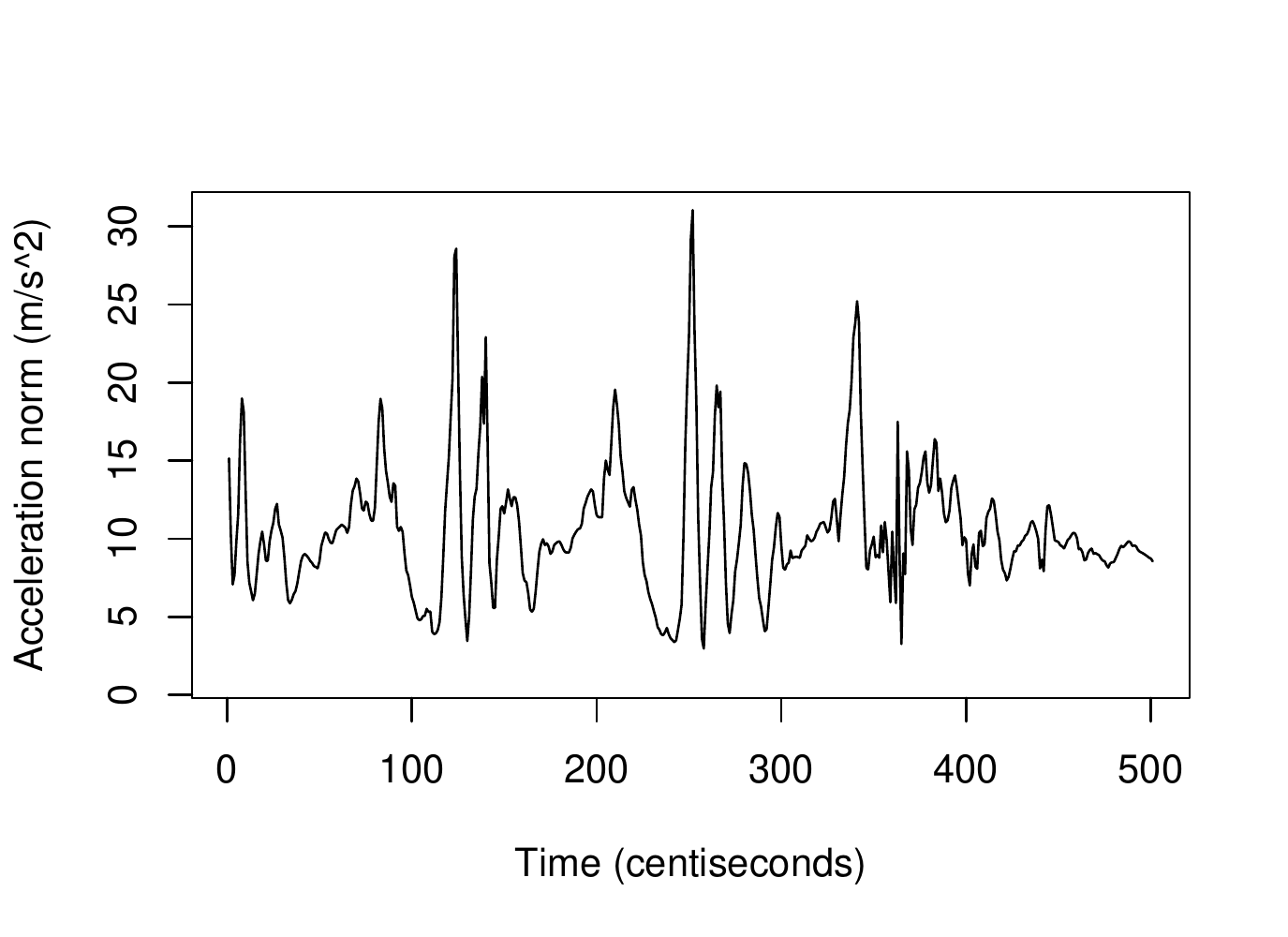}} 	
\caption{Norm of the acceleration experienced by a mobile phone when its owner is walking (a), and not walking, but gesticulating with the mobile in his/her hand (b).} 
\label{fig:examples-acc}
\end{figure}

The task we have chosen to conduct our experiments is the detection of the walking activity on smartphones. 
It is relatively easy to detect the walking activity and even count the steps when a person walks ideally with the mobile in the palm of his/her hand, facing upwards and without moving it too much. However, the situation gets much worse in real life, when the orientation of the mobile with respect to the body, as well as its location (hand, bag, pocket, ear, etc.), may change constantly as the person moves~\cite{casado2020,rodriguez2018}. 
Figure~\ref{fig:examples-acc} shows the complexity of this problem with a real example. In this figure we can see the norm  of the acceleration experienced by a mobile phone while its owner is doing two different activities. The person and the device are the same in both cases. In Figure~\ref{fig:steps-example}, we can see the signal obtained when the person is walking with the mobile in the pocket. Figure~\ref{fig:noise-example} shows a very similar acceleration signal experienced by the mobile, but in this case when the user is standing still with the mobile in his hand, without walking, but gesticulating with the arms in a natural way. 

In order to evaluate the performance of our method and the evolution of the models over the time, we require an independent test set that is representative enough of the problem. Thus, we started by building a fully labeled test dataset, for which 77 different people participated. We asked the volunteers to perform different walking and non-walking activities in several conditions. Since manual data labeling would be very time-consuming, volunteers carried a set of sensorized devices in their legs (besides their own mobile) tied with sports armbands in order to automatically get a ground truth~\cite{casado2020,rodriguez2018}.  Several features (21) from time and frequency domains were extracted from the 9-dimensional time series composed by the 3 axis of each of the 3 inertial sensors of the smartphone: accelerometer, gyroscope and magnetometer. This fully labeled dataset is composed of 7917 instances when dividing the data in time windows of 2.5 seconds. The ultimate goal of this set is to help us to evaluate the performance of the models we obtain applying \ourAcronym{}. Nevertheless, we decided to apply the traditional learning approach first in order to have a baseline. Remember that in the context of distributed devices it is not always feasible to apply traditional methods. However, for this task it is, because we were able to obtain a sufficient amount of well-labeled data from a high number of users. 
This is one of the main reasons why we chose this problem to test our proposal. Thus, we applied some of the most popular and widely known supervised classification methods on this dataset:
\begin{itemize}[noitemsep,topsep=0pt,parsep=0pt,partopsep=0pt]
    \item a simple Na{\"\i}ve Bayes (NB),
    \item a Generalized Linear Model (GLM), 
    \item a C5.0 Decision Tree (C5.0),
    \item a Support Vector Machine with radial basis function kernel~(RBF SVM),
    \item a Random Forests~(RF),
    \item a Stochastic Gradient Boosting (SGB).
\end{itemize}
Table~\ref{tab:old-work} summarize the most relevant results we achieved. Optimal hyperparameters for each of these traditional classifiers were estimated applying 10-fold cross validation on the dataset. We also tried several Convolutional Neural Network~(CNN) architectures~\cite{casado2020}. Table~\ref{tab:old-work} shows the results provided by the best one. Note that in this last case, as the size of the dataset is too small for deep learning, we had to to carry out an oversampling process to generate more instances artificially. 
The training and testing of the different models was carried out using the framework provided by the R language. In particular, for the training of the traditional models (Na{\"\i}ve Bayes, GLM, C5.0, etc.) we used the implementations already provided by the \texttt{caret} package~\cite{caretpkg}. In the case of the CNNs, we employed the \texttt{keras} package~\cite{keraspkg}. 

\begin{table}[htbp]
\centering
\caption{Performance of several supervised classifiers, trained and tested on a dataset of 77 different people~\cite{casado2020}.}
\label{tab:old-work}
\begin{threeparttable}
\resizebox{\linewidth}{!}{%
\begin{tabular}{l|c|c|c|}
	\cline{2-4}
	& \multicolumn{1}{c|}{\textbf{Balanced Accuracy}} & \multicolumn{1}{c|}{\textbf{Sensitivity}} & \multicolumn{1}{c|}{\textbf{Specificity}} \\ \hline
	\multicolumn{1}{|l|}{\textbf{Na{\"\i}ve Bayes}} & 0.8938 & 0.9580 & 0.8295 \\ \hline
	\multicolumn{1}{|l|}{\textbf{GLM}\tnote{1}} & 0.8855  & 0.8826 & 0.8884  \\ \hline
	\multicolumn{1}{|l|}{\textbf{C5.0 Tree}} & 0.9108 & 0.9015 & 0.9200  \\ \hline
	\multicolumn{1}{|l|}{\textbf{RBF SVM}\tnote{2}} & 0.9368 & 0.9410 & 0.9326 \\ \hline
	\multicolumn{1}{|l|}{\textbf{RF}\tnote{3}} & 0.9444 & 0.9375 & 0.9516 \\ \hline
	\multicolumn{1}{|l|}{\textbf{SGB}\tnote{4}} & 0.9324  & 0.9406 & 0.9242  \\ \hline
	\multicolumn{1}{|l|}{\textbf{CNN}\tnote{3}} & 0.9791 & 0.9755 & 0.9822 \\ \hline
\end{tabular}%
}
\begin{tablenotes}
      \footnotesize
      \item[1] GLM = Generalized Linear Model
      \item[2] RBF SVM = Support Vector Machine with radial basis kernel
      \item[3] RF = Random Forests
      \item[4] SGB = Stochastic Gradient Boosting
      \item[5] CNN = Convolutional Neural Network
    \end{tablenotes}
    \end{threeparttable}
\end{table}

However, despite the high performances shown in Table~\ref{tab:old-work}, we must admit that these numbers are unrealistic. Obtaining performances like those involves months of work, collecting and labeling data from different volunteers and re-tuning and re-obtaining increasingly robust models until significant improvements are no longer achieved. It is at that moment when the model is finally put into exploitation. In fact, we have tried all the models in Table~\ref{tab:old-work} on a smartphone and we have identified several situations, quite common in everyday life, where all of them fail, such as walking with the mobile in a back pocket of the pants, carrying the mobile in a backpack or shaking it slowly in the hand simulating the forces the device experience when the user is walking. No matter how complete we believe our dataset is, in many real problems there are going to be some situations that are poorly represented in the training data. The classical solution to this would be to collect new data of these situations where the model fails, try to obtain a better model robust to this situations and, once achieved, replace the old model with the new one. 

In order to build an appropriate training dataset to evaluate the federated, continual and semi-supervised learning possibilities of our proposal, we developed an Android application that samples and logs the inertial data on mobile phones continuously, after being processed. These data is all unlabeled. Nevertheless, the app allows the user to indicate whether he/she is walking or not through a switch button in the graphical interface, but this is optional, so depending on the user's willingness to participate, there will be more or less labeled data. The app also labels autonomously some examples applying a series of heuristic rules when it comes to clearly identifiable positives or negatives (e.g., when the phone is at rest). With this app, we collected partially labeled data from 10 different people. Participants installed our application and recorded data continuously while they were performing their usual routine. 
Table~\ref{tab:summary-data} summarizes the data distribution in the new semi-labeled dataset by user, and compares it with the already presented test set.

\begin{table}[htbp]
\centering
\caption{Summary of the training and test sets, indicating the number of examples attending to the label.}
\label{tab:summary-data}
\resizebox{\linewidth}{!}{%
\begin{tabular}{cl|c|c|c|c|}
	\cline{3-6}
	& & \textbf{Walking} & \textbf{Not walking} & \textbf{Unlabeled} & \textbf{Total} \\ \hline
	\multicolumn{1}{|c|}{\multirow{10}{*}{\textbf{Training}}} & \multicolumn{1}{|l|}{\textbf{user 1}} & 3130 & 2250 & 4620 & 10000 \\ \cline{2-6} 
	\multicolumn{1}{|c|}{\multirow{10}{*}{\textbf{set}}}   & \multicolumn{1}{|l|}{\textbf{user 2}} & 2519 & 4359 & 3122 & 10000 \\ \cline{2-6} 
	\multicolumn{1}{|c|}{} & \multicolumn{1}{|l|}{\textbf{user 3}} & 186  &  325 &  125 & 636   \\ \cline{2-6} 
	\multicolumn{1}{|c|}{} & \multicolumn{1}{|l|}{\textbf{user 4}} & 2432 & 2455 & 5113 & 10000 \\ \cline{2-6} 
	\multicolumn{1}{|c|}{} & \multicolumn{1}{|l|}{\textbf{user 5}} & 233  & 2785 & 6982 & 10000 \\ \cline{2-6} 
	\multicolumn{1}{|c|}{} & \multicolumn{1}{|l|}{\textbf{user 6}} & 554  & 1821 &  69  & 2444  \\ \cline{2-6} 
	\multicolumn{1}{|c|}{} & \multicolumn{1}{|l|}{\textbf{user 7}} & 2582 & 2052 & 769  & 5403  \\ \cline{2-6} 
	\multicolumn{1}{|c|}{} & \multicolumn{1}{|l|}{\textbf{user 8}} & 232  & 678  & 229  & 1139  \\ \cline{2-6} 
	\multicolumn{1}{|c|}{} & \multicolumn{1}{|l|}{\textbf{user 9}} & 1151 & 2669 & 6180 & 10000 \\ \cline{2-6} 
	\multicolumn{1}{|c|}{} & \multicolumn{1}{|l|}{\textbf{user 10}}& 2329 & 2669 & 5002 & 10000 \\ \cline{2-6} 
	\multicolumn{1}{|c|}{} & \multicolumn{1}{|l|}{\textbf{Total}}  &15348 &22063 &32211 & 69622 \\ \hline \hline
    \multicolumn{1}{|c|}{\textbf{Test set}} & \multicolumn{1}{|l|}{\textbf{Total}}& 6331 & 1586 & 0    & 7917  \\ \hline
\end{tabular}%
}
\end{table}

Applying the same techniques of Table~\ref{tab:old-work} but now using the new dataset for training and the old one for testing (Table~\ref{tab:summary-data}), the results obtained are those of Table~\ref{tab:old-work-new-test}. If we compare both tables, Table~\ref{tab:old-work} and Table~\ref{tab:old-work-new-test}, it is clear that the latter results are substantially worse---approximately 10\% worse in all the cases---. This is logical, since training now is carried out using a totally new dataset, obtained in a more realistic and unconstrained way, which contains biased and partially labeled data from a much smaller number of users. 
Note that we have not obtained the new results using CNNs in Table~\ref{tab:old-work-new-test}. This is because, for the experiments in Table~\ref{tab:old-work}, a dedicated mobile device was used, so recording raw data for training a CNN was not a problem. However, the new training set was obtained in a federated context using the participants' own smartphones, so recording raw data was not an option.

\begin{table}[htbp]
\centering
\caption{Performance of supervised classifiers, trained and tested using the datasets described in Table~\ref{tab:summary-data}.}
\label{tab:old-work-new-test}
\resizebox{\linewidth}{!}{%
\begin{tabular}{l|c|c|c|}
	\cline{2-4}
	& \multicolumn{1}{c|}{\textbf{Balanced Accuracy}} & \multicolumn{1}{c|}{\textbf{Sensitivity}} & \multicolumn{1}{c|}{\textbf{Specificity}} \\ \hline
	\multicolumn{1}{|l|}{\textbf{Na{\"\i}ve Bayes \quad}} & 0.7949 & 0.9044 & 0.6854 \\ \hline
	\multicolumn{1}{|l|}{\textbf{GLM}} & 0.7221  & 0.8206 & 0.6236  \\ \hline
	\multicolumn{1}{|l|}{\textbf{C5.0 Tree}} & 0.8173 & 0.8842 & 0.7503  \\ \hline
	\multicolumn{1}{|l|}{\textbf{RBF SVM}} & 0.8378 & 0.9430 & 0.7327 \\ \hline
	\multicolumn{1}{|l|}{\textbf{RF}} & 0.8546 & 0.9021 & 0.8071 \\ \hline
	\multicolumn{1}{|l|}{\textbf{SGB}} & 0.8596  & 0.9084 & 0.8108  \\ \hline
\end{tabular}%
}
\end{table}

\begin{figure*}[htb]
	\centering
	\subfloat[Using NB as base classifier.]{\label{fig:roc-normal-nb}\includegraphics[width=0.32\linewidth,trim={1.3cm 0cm 0.5cm 0.8cm},clip]{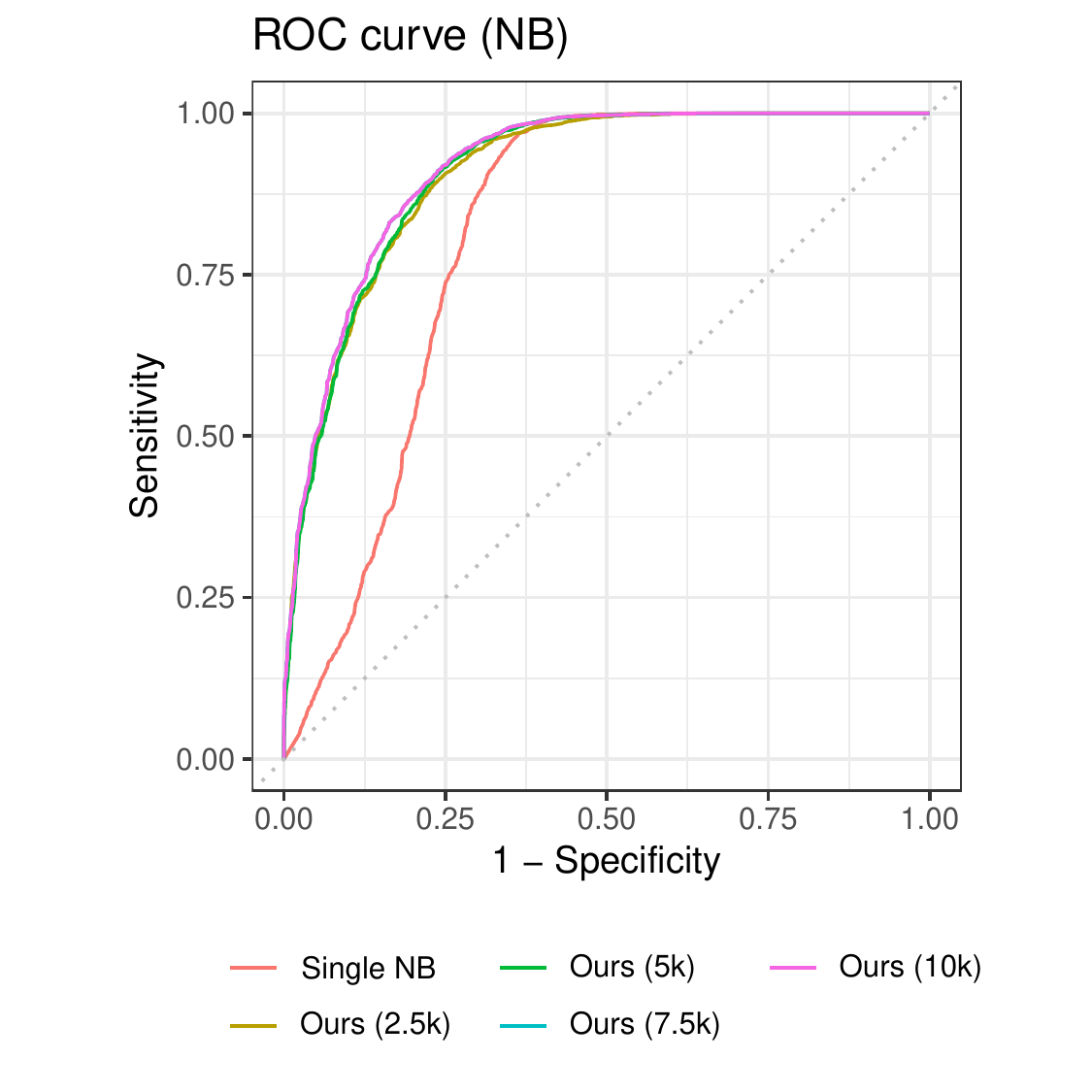}} 
	\subfloat[Using GLM as base classifier.]{\label{fig:roc-normal-glm}\includegraphics[width=0.32\linewidth,trim={1.3cm 0cm 0.5cm 0.8cm},clip]{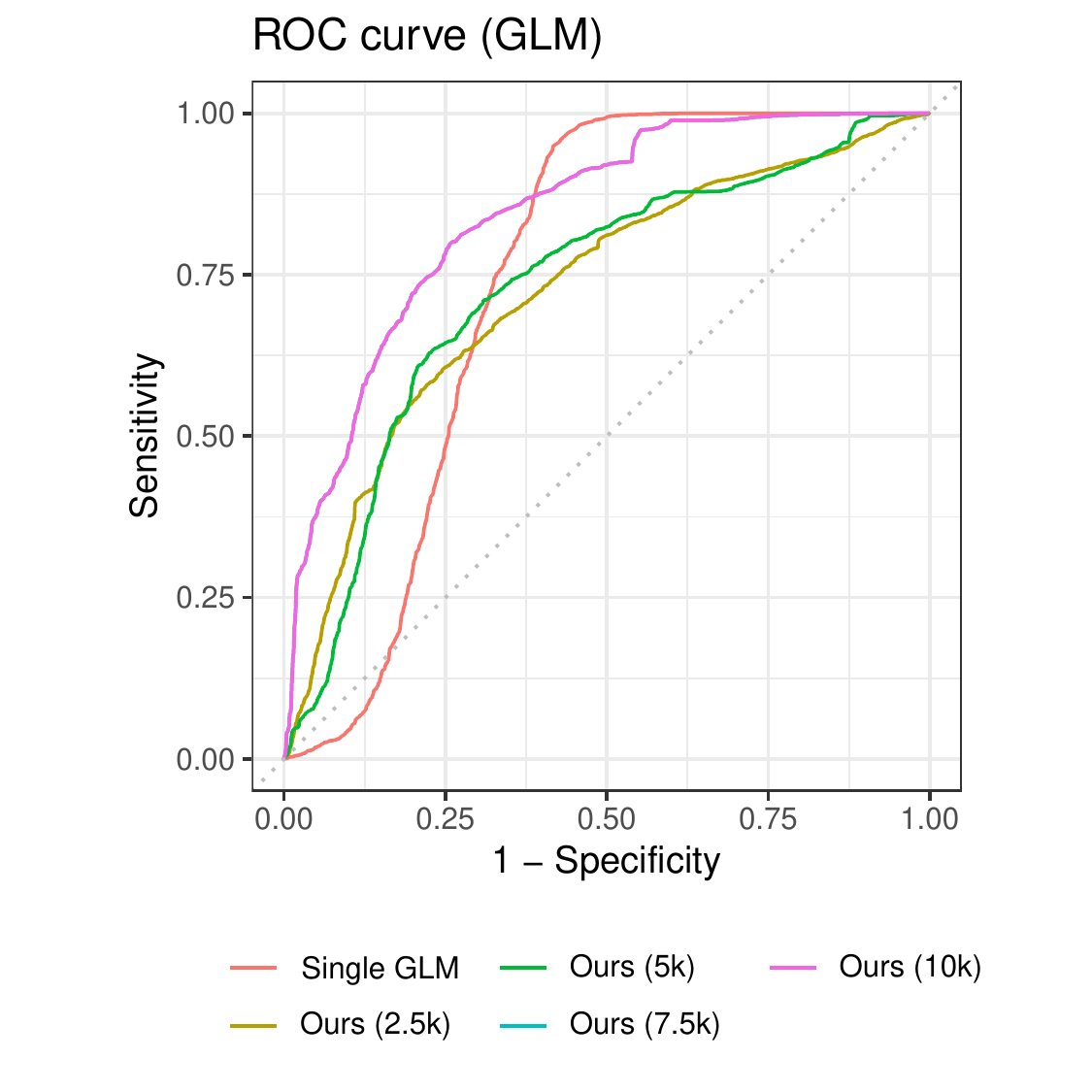}} 
	\subfloat[Using C5.0 as base classifier.]{\label{fig:roc-normal-c50}\includegraphics[width=0.32\linewidth,trim={1.3cm 0cm 0.5cm 0.8cm},clip]{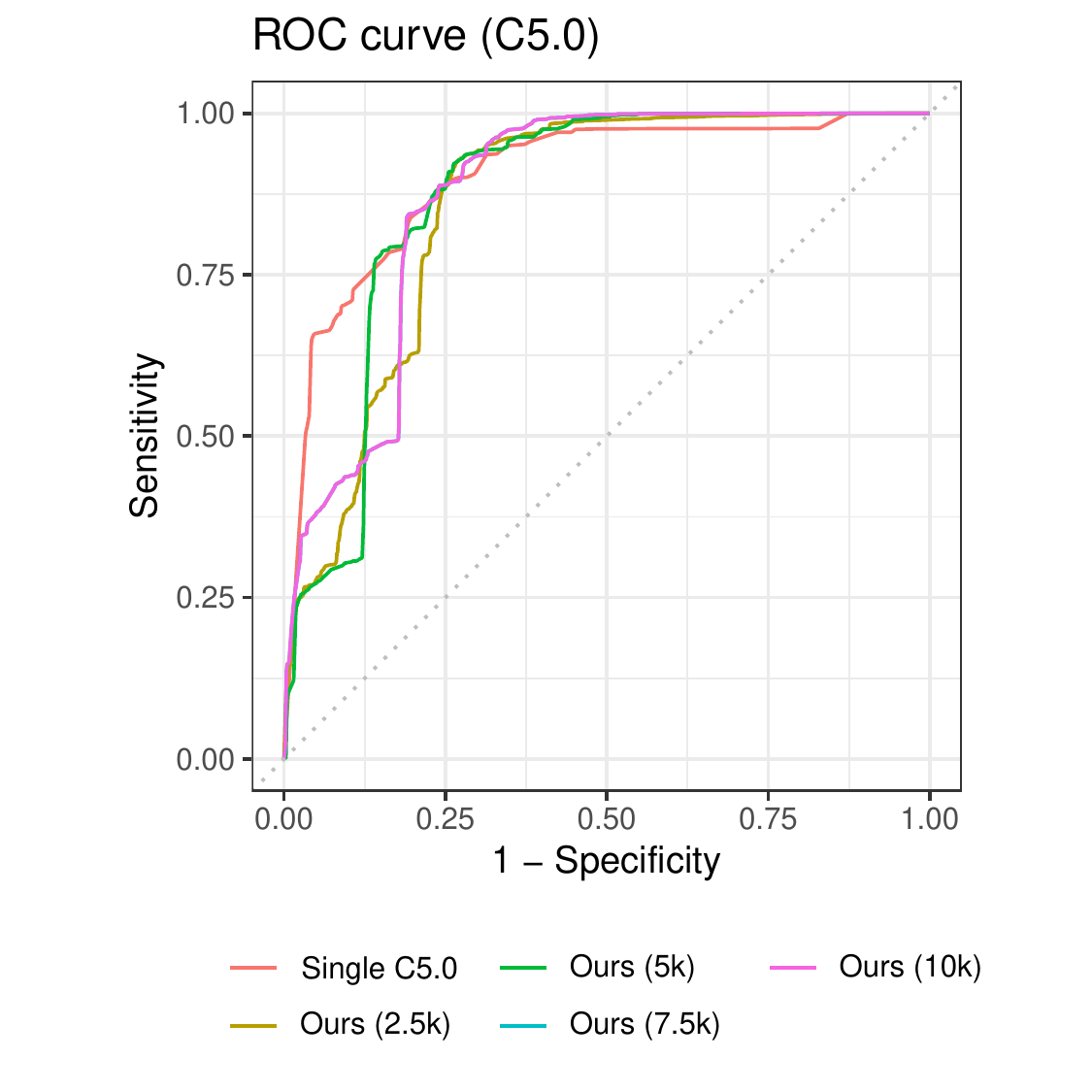}} \\ 
	\subfloat[Using RBF SVM as base classifier.]{\label{fig:roc-normal-svm}\includegraphics[width=0.32\linewidth,trim={1.3cm 0cm 0.5cm 0.8cm},clip]{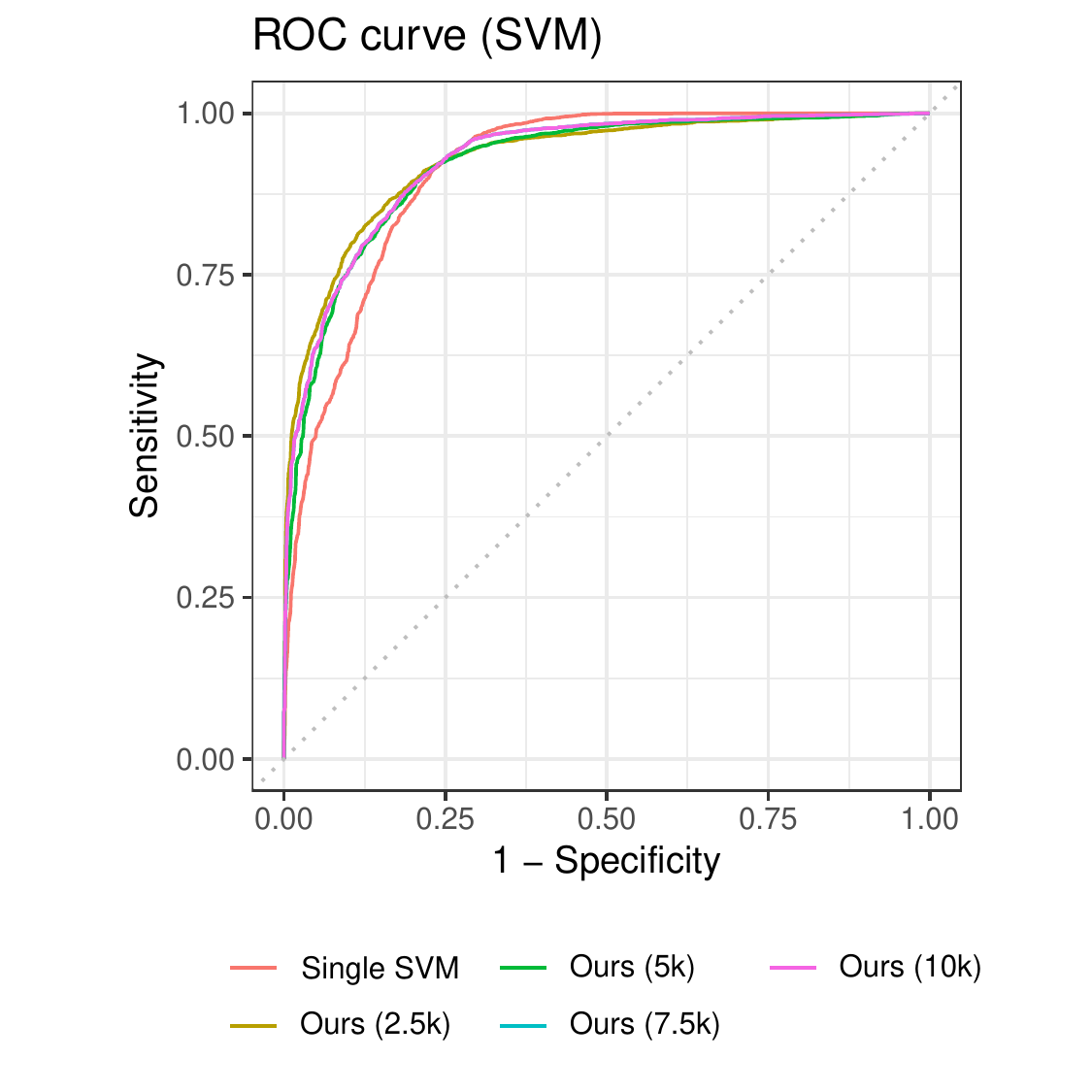}} 
	\subfloat[Using RF as base classifier.]{\label{fig:roc-normal-rf}\includegraphics[width=0.32\linewidth,trim={1.3cm 0cm 0.5cm 0.8cm},clip]{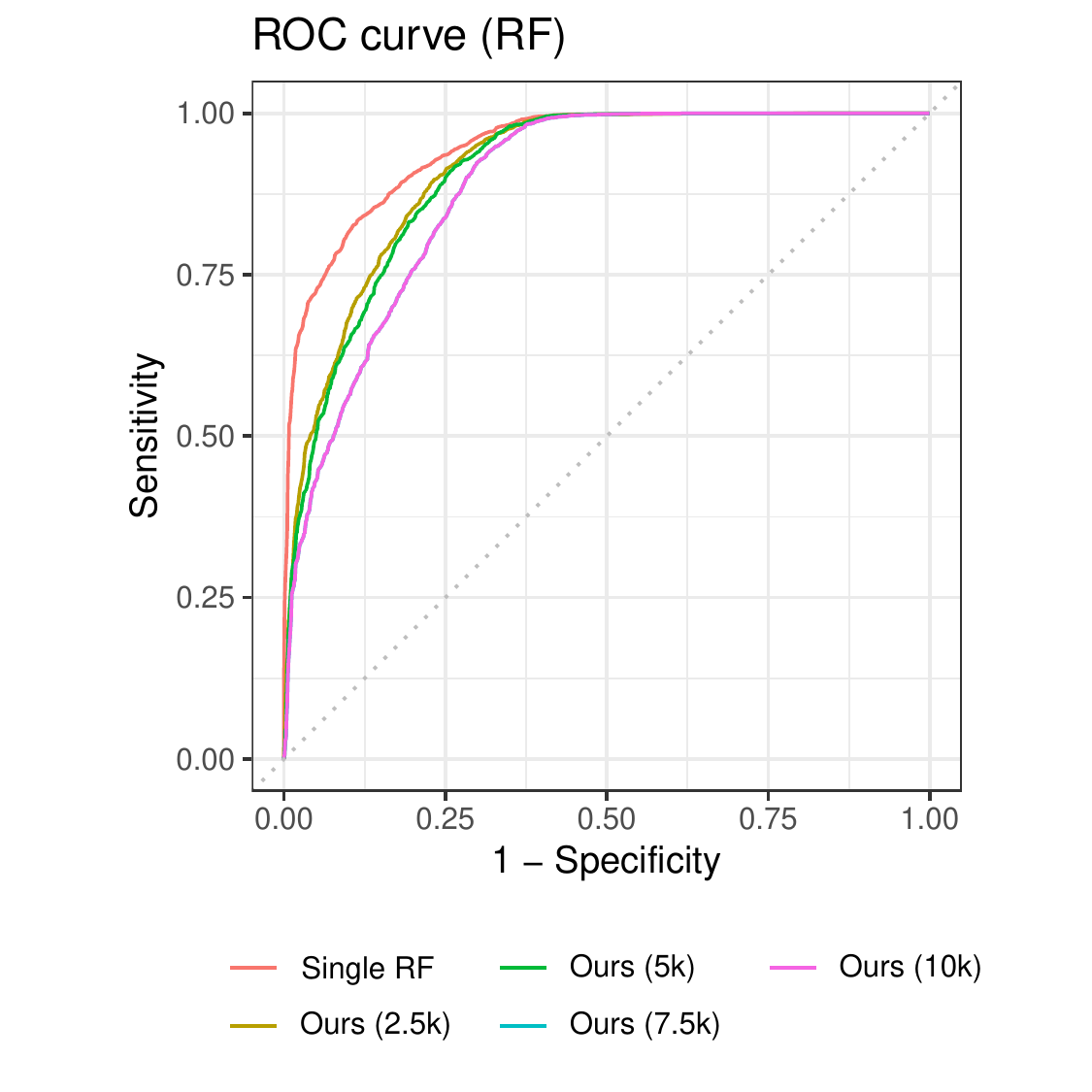}} 
	\subfloat[Using SGB as base classifier.]{\label{fig:roc-normal-sgb}\includegraphics[width=0.32\linewidth,trim={1.3cm 0cm 0.5cm 0.8cm},clip]{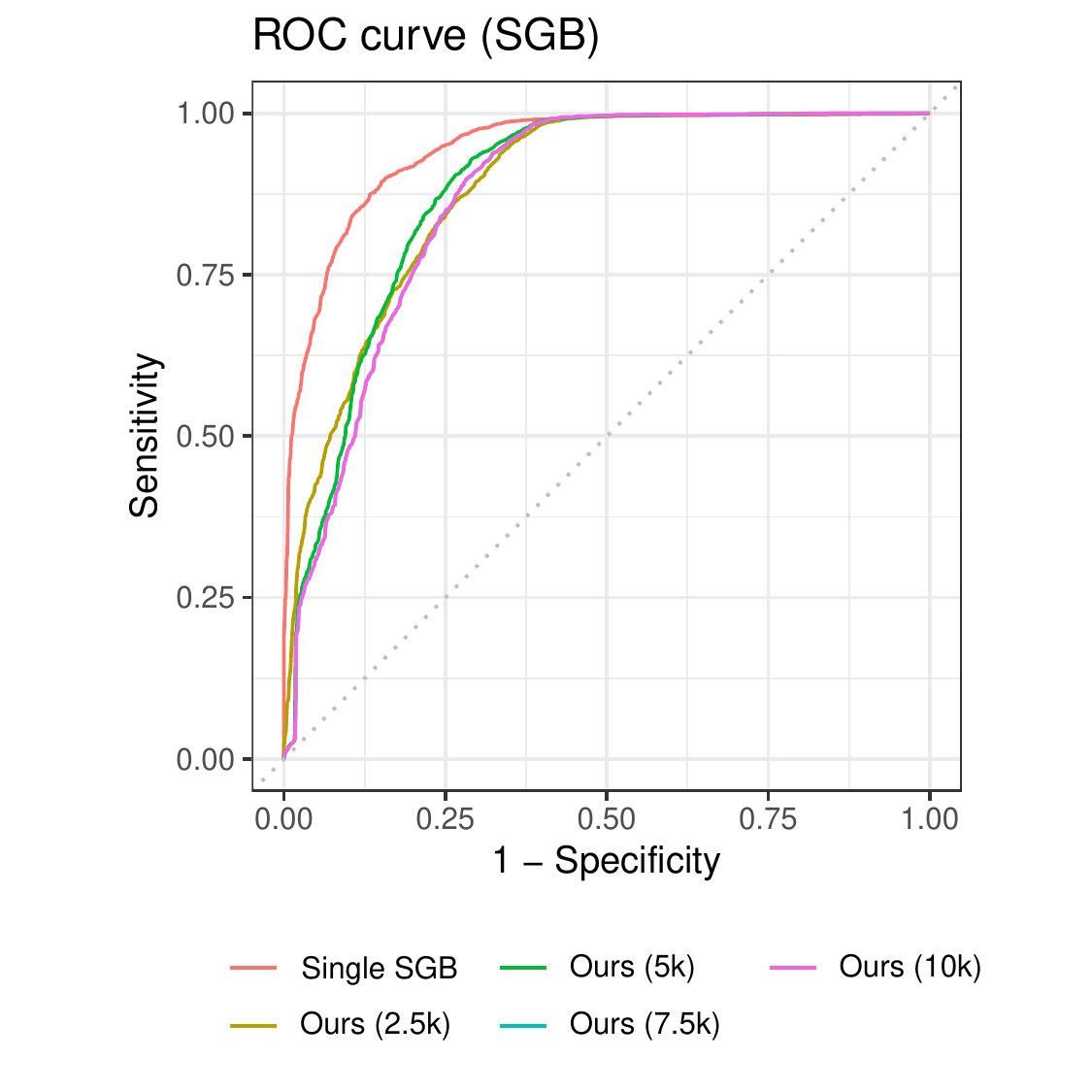}} 
	\caption{ROC curves provided by a single classifier (Table~\ref{tab:old-work-new-test}) compared with the global model from the federated training (Table~\ref{tab:method-normal}) at iterations 2500, 5000, 7500 and 10000.}
	\label{fig:roc-normal}
\end{figure*}

Using \ourAcronym{}, the training data from Table~\ref{tab:summary-data} is distributed among the different local devices. Data acquisition is continuous over time. For simplicity, we assume all local devices record data with the same frequency and, therefore, each iteration of \ourAcronym{} will correspond to a new pattern. It is important to remember that in our proposal each local model is an ensemble that is updated over time and has a maximum size, $M_l$. Similarly, a subset of local models of size $M_g$ is selected on the server to form the global ensemble model. In this experiments we used ensembles of 5 models, both locally and globally, i.e., $M_l = 5$ and $M_g = 5$.
Table~\ref{tab:method-normal} shows the performance of the global model after 10000 iterations using each of the methods in Table~\ref{tab:old-work-new-test} as base local learners. Figure~\ref{fig:roc-normal} compares the Receiver Operating Characteristic (ROC) curve of the traditional models from Table~\ref{tab:old-work-new-test} with its federated equivalent at different iterations. 

\begin{table}[htbp]
\centering
\caption{Performance of \ourAcronym{} after 10000 iterations, when $M_g = M_l = 5$, using as base classifier the methods in Table~\ref{tab:old-work-new-test}.}
\label{tab:method-normal}
\resizebox{\linewidth}{!}{%
\begin{tabular}{l|c|c|c|}
	\cline{2-4}
	& \multicolumn{1}{c|}{\textbf{Balanced Accuracy}} & \multicolumn{1}{c|}{\textbf{Sensitivity}} & \multicolumn{1}{c|}{\textbf{Specificity}} \\ \hline
	\multicolumn{1}{|l|}{\textbf{\ourAcronym{} + NB}} & 0.8280 & 0.7909 & 0.8651 \\ \hline
	\multicolumn{1}{|l|}{\textbf{\ourAcronym{} + GLM}} & 0.7405  & 0.8668 & 0.6141  \\ \hline
	\multicolumn{1}{|l|}{\textbf{\ourAcronym{} + C5.0}} & 0.8168 & 0.9665 & 0.6671  \\ \hline
	\multicolumn{1}{|l|}{\textbf{\ourAcronym{} + SVM}} & 0.8663 & 0.8499 & 0.8827 \\ \hline
	\multicolumn{1}{|l|}{\textbf{\ourAcronym{} + RF}} & 0.8314 & 0.9489 & 0.7129 \\ \hline
	\multicolumn{1}{|l|}{\textbf{\ourAcronym{} + SGB}} & 0.8231  & 0.9187 & 0.7264  \\ \hline
\end{tabular}%
}
\end{table}

As we can see, our federated proposal gives similar performances to those achieved by a single model trained with all data available. However, the comparison between Tables~\ref{tab:old-work-new-test} and~\ref{tab:method-normal} is not fair, as they correspond to two very different approaches. Note that in the \ourAcronym{} setting of Table~\ref{tab:method-normal} we are only using half of the users to build the global ensemble model ($M_g = 5$), whereas the models in Table~\ref{tab:old-work-new-test} use all labeled data from all users. Furthermore, the use of RF or SGB as base classifiers does not seem to be the best option in the federated scenario, since these algorithms are already ensembles. For these cases there are probably much more efficient combination strategies that could be tested. For example, by creating an ensemble of Random Forests we are combining the final decisions of each of the forest, when it might be better to combine the decisions of all the trees of each forest, at a lower level. Proposals like this are out of the scope of this work, in which we have opted for a global learning method scalable to multiple settings.

\begin{figure}[htb]
\centering
\includegraphics[width=0.9\linewidth,trim={0 2cm 0 0},clip]{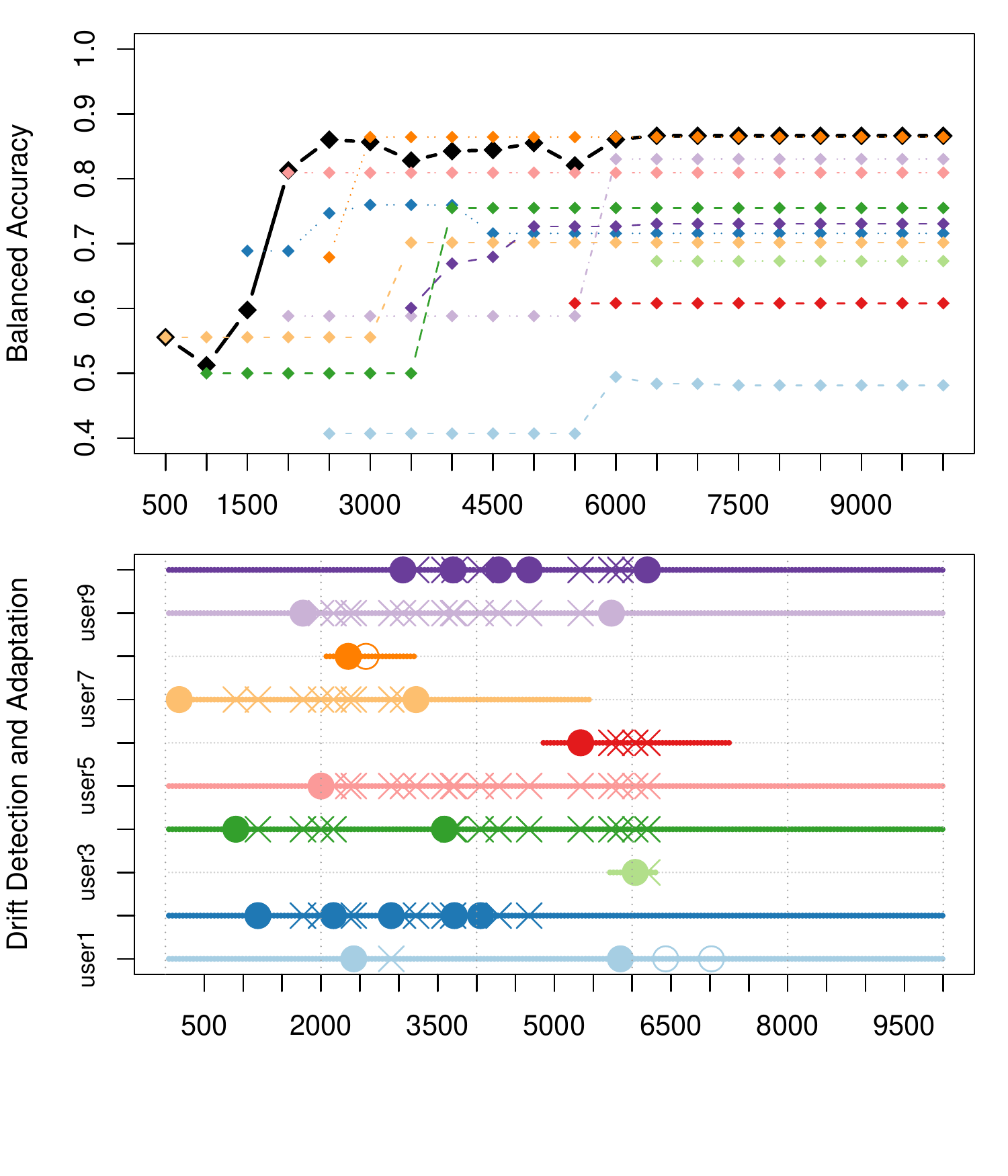}
\caption{Results for \ourAcronym{} using the RBF SVM as local base classifier and $M_g = 5$ and $M_l = 5$. The upper graph shows the evolution of the accuracy over time. The bottom graph shows the drifts. The thick black line corresponds to the global model. The rest of the coloured lines are each of the 10 anonymous users.}
\label{fig:results-method-5mg-svm}
\end{figure}

In order to better illustrate the learning process of our method, we will take as an example the case where we use a RBF SVM as base classifier (fourth row in  Table~\ref{tab:method-normal}). Figure~\ref{fig:results-method-5mg-svm} and Table~\ref{tab:results-method-5mg-svm} show how the performance of \ourAcronym{} evolves over the iterations. 
The upper graph in Figure~\ref{fig:results-method-5mg-svm} shows the evolution of the accuracies of all the models ---from each of the 10 users and also the global one---. Table~\ref{tab:results-method-5mg-svm} provides the exact values of these accuracies.  In each column of Table~\ref{tab:results-method-5mg-svm}, those local models that are part of the global model in that iteration are marked with an asterisk~(*). We can appreciate that the global ensemble is not always selecting the most accurate local models. This actually makes sense, since local models are chosen using the Distributed Effective Voting~(DEV) from Figure~\ref{fig:effective-voting}, which is based on the voting carried out by a subset of $q$ users randomly chosen (5 in this case, because $q = M_g = 5$). The fact that some suboptimal local models are being selected indicates that there are devices that value these models highly, which means that the selection criteria of the voters can be improved. Remember that \ourAcronym{} needs a mechanism for selecting local models to build the global one, but this could be other than DEV. However, designing a good selection method is not straightforward as some challenges must be addressed. Firstly, local models must be evaluated without centralized data being available anywhere and without even being able to retain all local data. In addition, a balance is required between precision in the selection and efficiency in the overall system. Improving the voting system may involve more communication between server and devices, which limits the scalability of the proposal. Nevertheless, in the results we have obtained we can notice how the global ensemble is able to provide good performance almost from the beginning. This leads us to think that perhaps it is not necessarily bad that some suboptimal classifiers are chosen and even that the diversity on the global ensemble is more important than the accuracy of each of its members.

\begin{table}[htb]
\centering
\caption{Detail of the balanced accuracies achieved in the \ourAcronym{} setting of Figure~\ref{fig:results-method-5mg-svm}.}
\label{tab:results-method-5mg-svm}
\begin{threeparttable}
\begin{tabular}{c|c|c|c|c|c|c|c|}
	\cline{2-5}
	\multicolumn{1}{l|}{} & \multicolumn{4}{c|}{\textbf{Number of iterations}} \\ \hline
	\multicolumn{1}{|c|}{\textbf{Model}} & \textbf{2500} & \textbf{5000} & \textbf{7500} & \textbf{10000} \\ \hline
	\multicolumn{1}{|c|}{\textbf{user 1}} & 0.4069* & 0.4069  & 0.4813  & 0.4813  \\ \hline
	\multicolumn{1}{|c|}{\textbf{user 2}} & 0.7468* & 0.7158* & 0.7158  & 0.7157  \\ \hline	\multicolumn{1}{|c|}{\textbf{user 3}} &   -     &   -     & 0.6730* & 0.6730*   \\ \hline
	\multicolumn{1}{|c|}{\textbf{user 4}} & 0.4999  & 0.7549* & 0.7549* & 0.7549*  \\ \hline
	\multicolumn{1}{|c|}{\textbf{user 5}} & 0.8093* & 0.8093* & 0.8093* & 0.8093*   \\ \hline
	\multicolumn{1}{|c|}{\textbf{user 6}} &   -     &   -     & 0.6080* & 0.6080*   \\ \hline
	\multicolumn{1}{|c|}{\textbf{user 7}} & 0.5555* & 0.7016  & 0.7016  & 0.7016   \\ \hline
	\multicolumn{1}{|c|}{\textbf{user 8}} & 0.6789  & 0.8644  & 0.8644  & 0.8644   \\ \hline
	\multicolumn{1}{|c|}{\textbf{user 9}} & 0.5882* & 0.5882* & 0.8304  & 0.8304  \\ \hline
	\multicolumn{1}{|c|}{\textbf{user 10}}&   -     & 0.7266* & 0.7306* & 0.7306*  \\ \hline \hline
	\multicolumn{1}{|c|}{\textbf{global}} & 0.8603  & 0.8552  & 0.8663  & 0.8663   \\ \hline
\end{tabular}%
\begin{tablenotes}
      \footnotesize
      \item * Cells marked with an asterisk indicate that the current model of that user was chosen to participate in the global ensemble.
\end{tablenotes}
\end{threeparttable}
\end{table}

In Figure~\ref{fig:results-method-5mg-svm}, the thick black line corresponds to the global model accuracy while the rest of the coloured lines are each of the 10 anonymous users. As in each iteration the unlabeled local data is labeled with the most recent global model, the more unlabeled data the device has, the more it will be enriched by the global knowledge. The bottom graph shows the drift detection and adaptation. Once again, each coloured line corresponds to one user. In those places where the line is not drawn it means that the device is not taking data. A circumference~($\circ$) on the line indicates when a drift has been detected and the local model has been updated. If the circumference is filled~($\bullet$) it indicates that the local model is chosen as one of the 5 models of the global ensemble --the other 4 chosen are marked with a cross ($\times$)--.

\begin{figure*}[htb]
	\centering
	\subfloat[Using NB as base classifier.]{\label{fig:roc-4nu-nb}\includegraphics[width=0.32\linewidth,trim={1.3cm 0cm 0.5cm 0.8cm},clip]{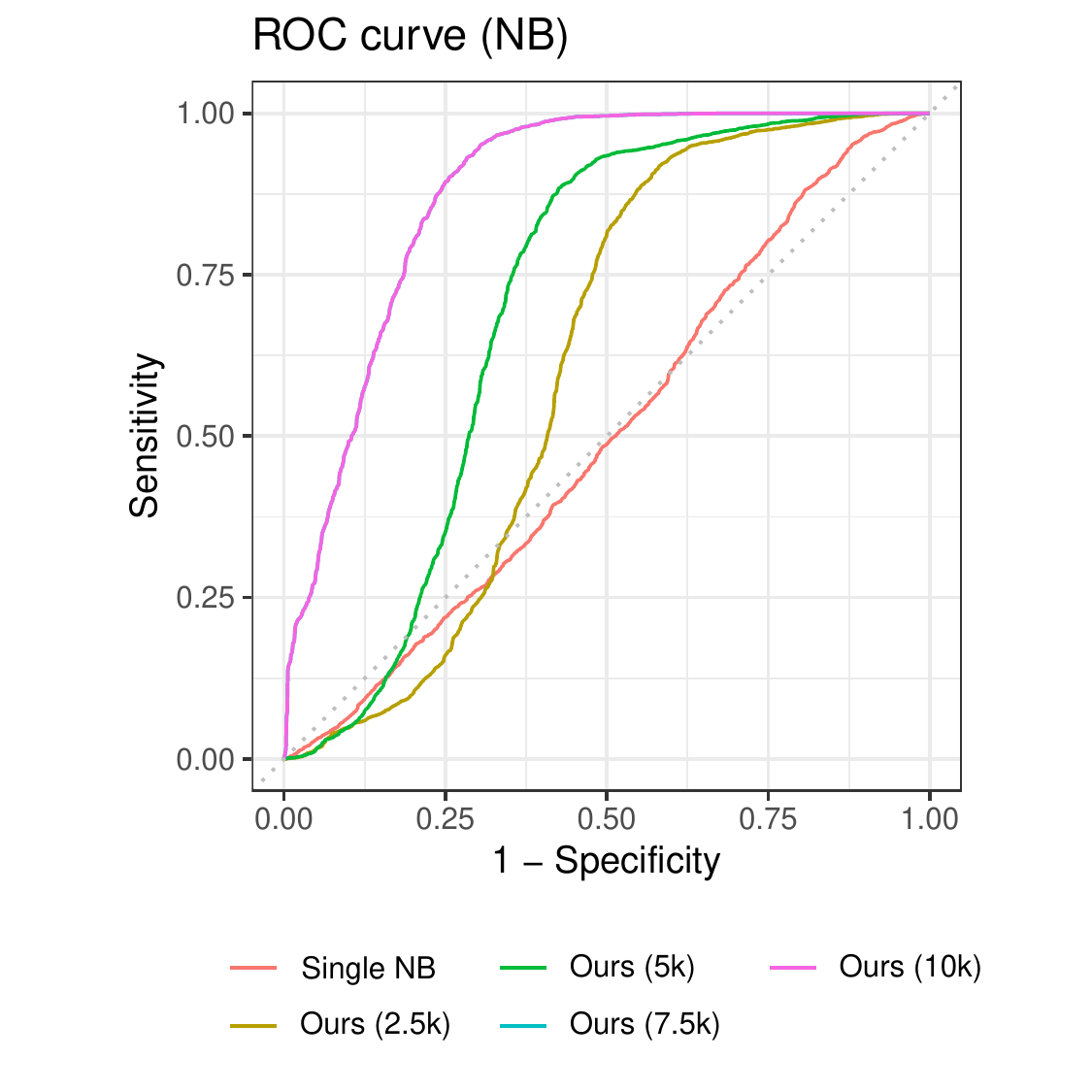}} 
	\subfloat[Using GLM as base classifier.]{\label{fig:roc-4nu-glm}\includegraphics[width=0.32\linewidth,trim={1.3cm 0cm 0.5cm 0.8cm},clip]{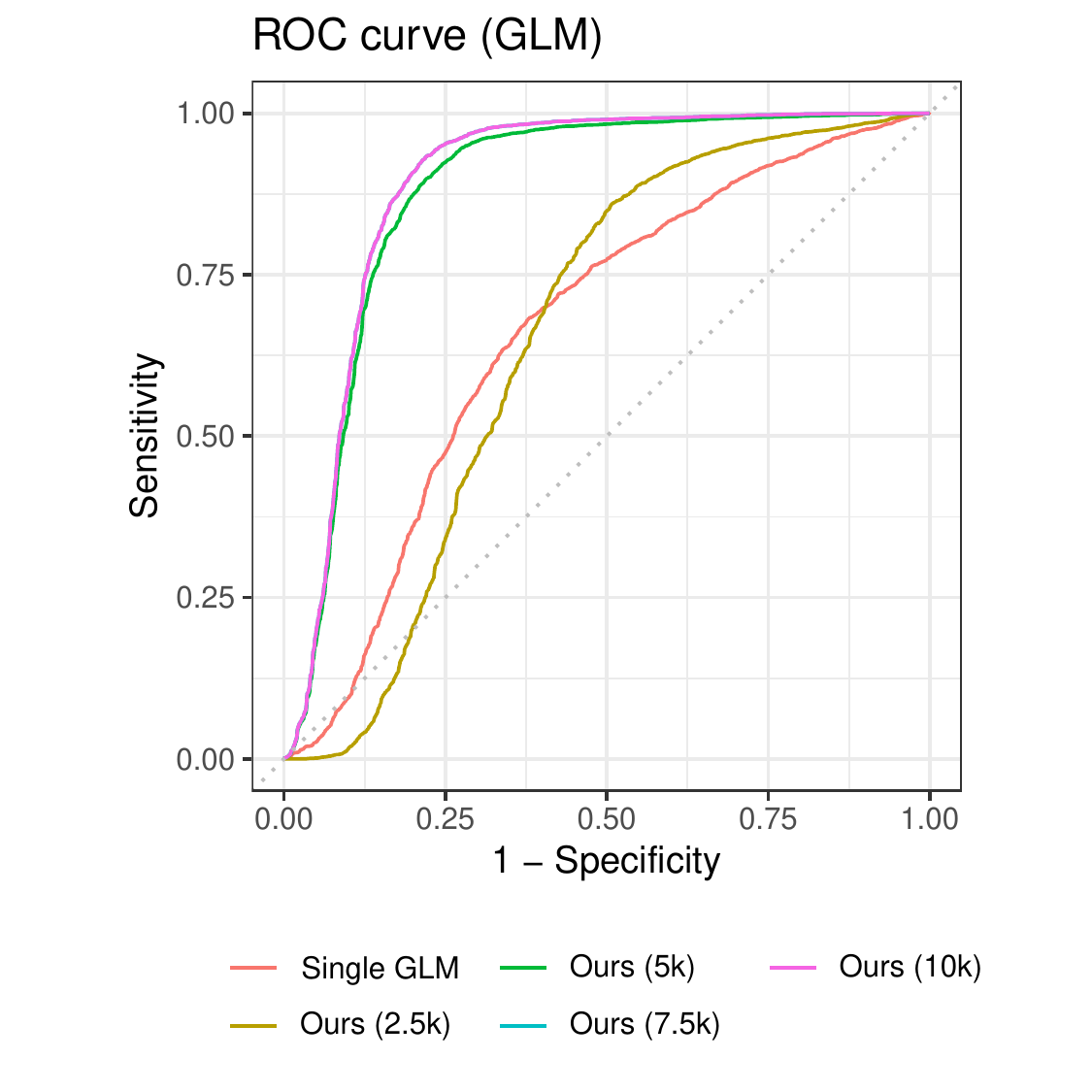}} 
	\subfloat[Using C5.0 as base classifier.]{\label{fig:roc-4nu-c50}\includegraphics[width=0.32\linewidth,trim={1.3cm 0cm 0.5cm 0.8cm},clip]{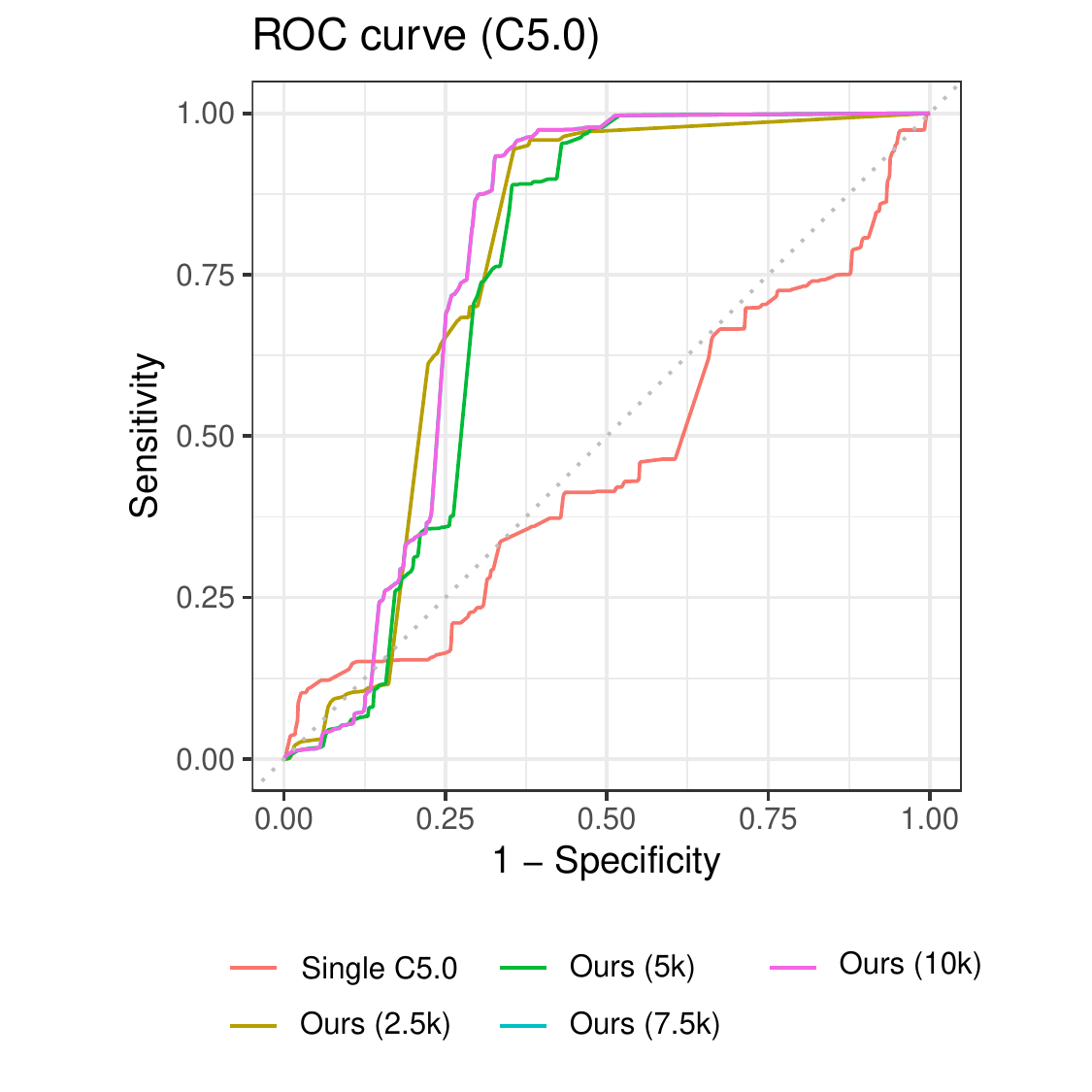}} \\ 
	\subfloat[Using RBF SVM as base classifier.]{\label{fig:roc-4nu-svm}\includegraphics[width=0.32\linewidth,trim={1.3cm 0cm 0.5cm 0.8cm},clip]{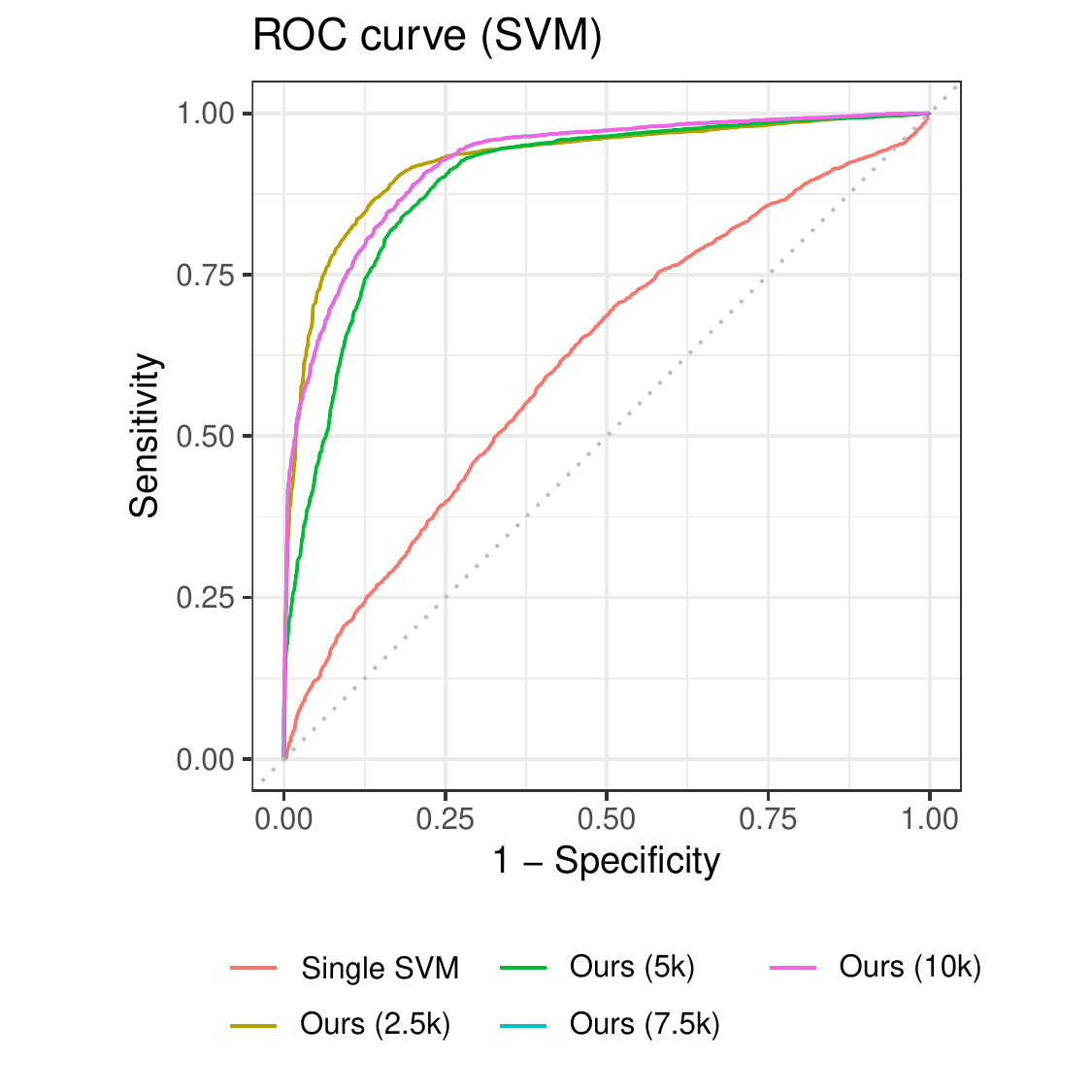}} 
	\subfloat[Using RF as base classifier.]{\label{fig:roc-4nu-rf}\includegraphics[width=0.32\linewidth,trim={1.3cm 0cm 0.5cm 0.8cm},clip]{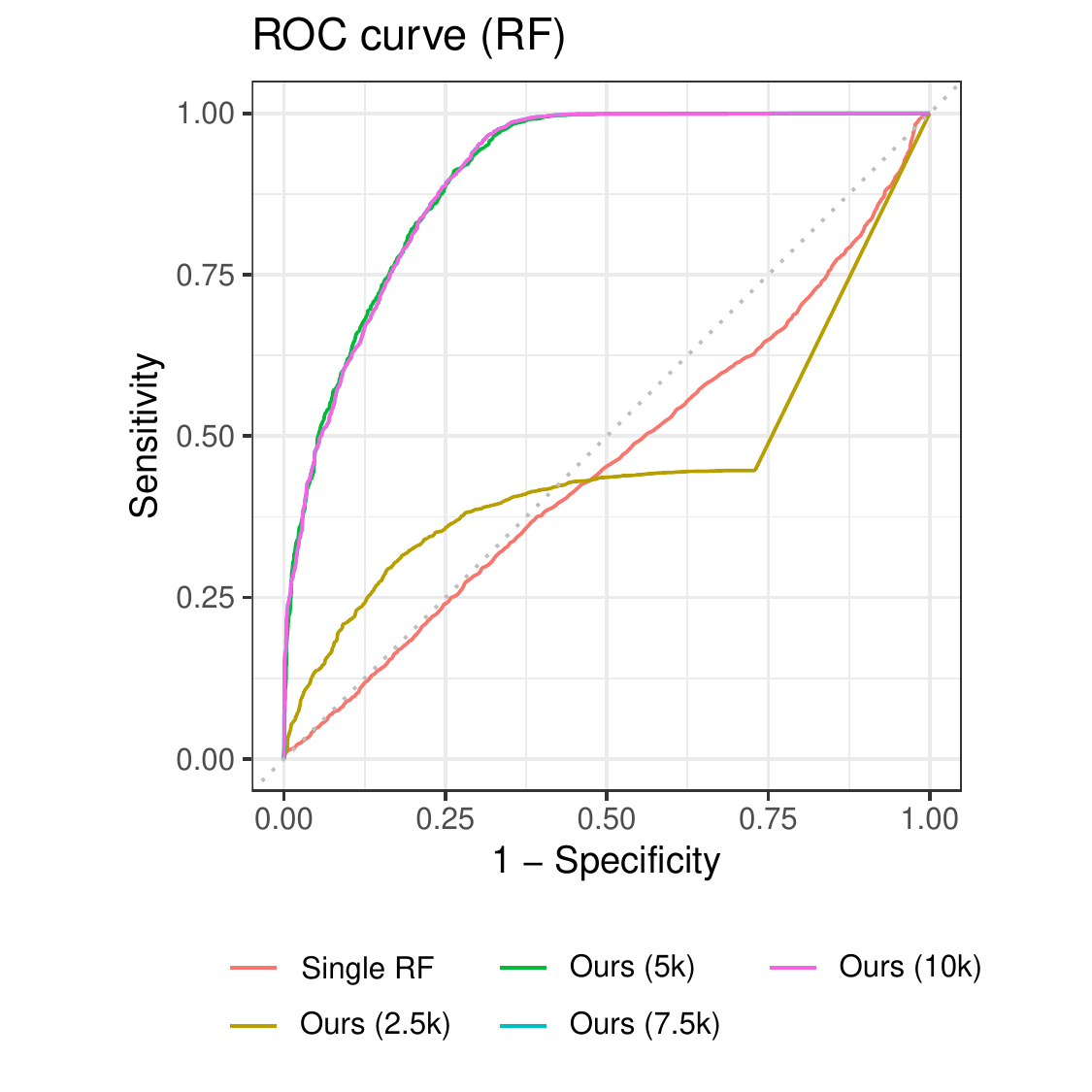}} 
	\subfloat[Using SGB as base classifier.]{\label{fig:roc-4nu-sgb}\includegraphics[width=0.32\linewidth,trim={1.3cm 0cm 0.5cm 0.8cm},clip]{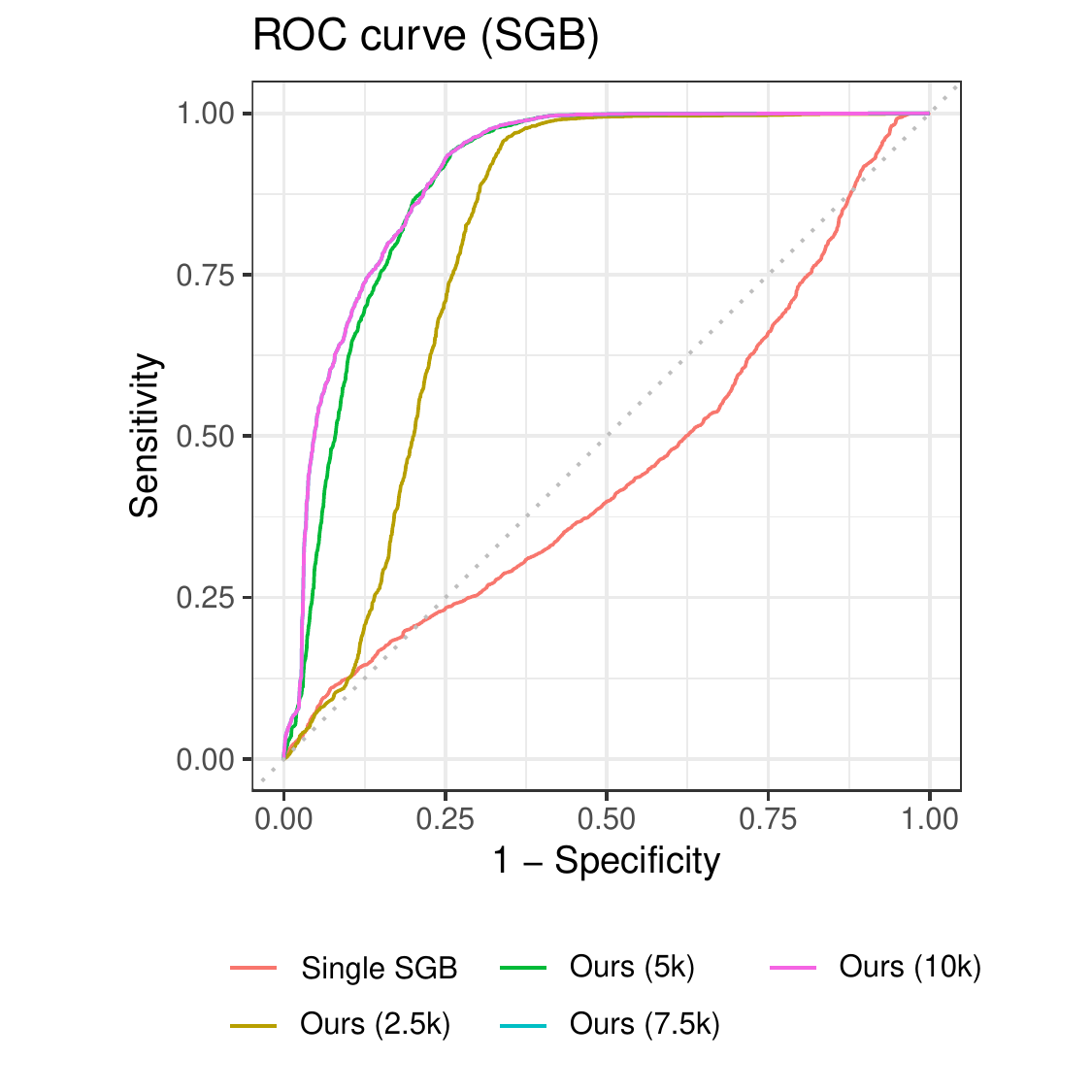}} 
	\caption{ROC curves provided by a single classifier (Table~\ref{tab:results-old-4noisyusers}) compared with the global model from the federated training (Table~\ref{tab:results-method-4noisyusers}) at iterations 2500, 5000, 7500 and 10000 when there are users who provide noisy data.}
	\label{fig:roc-4nu}
\end{figure*}

\begin{figure}[htb]
\centering
\includegraphics[width=0.9\linewidth,trim={0 2cm 0 0},clip]{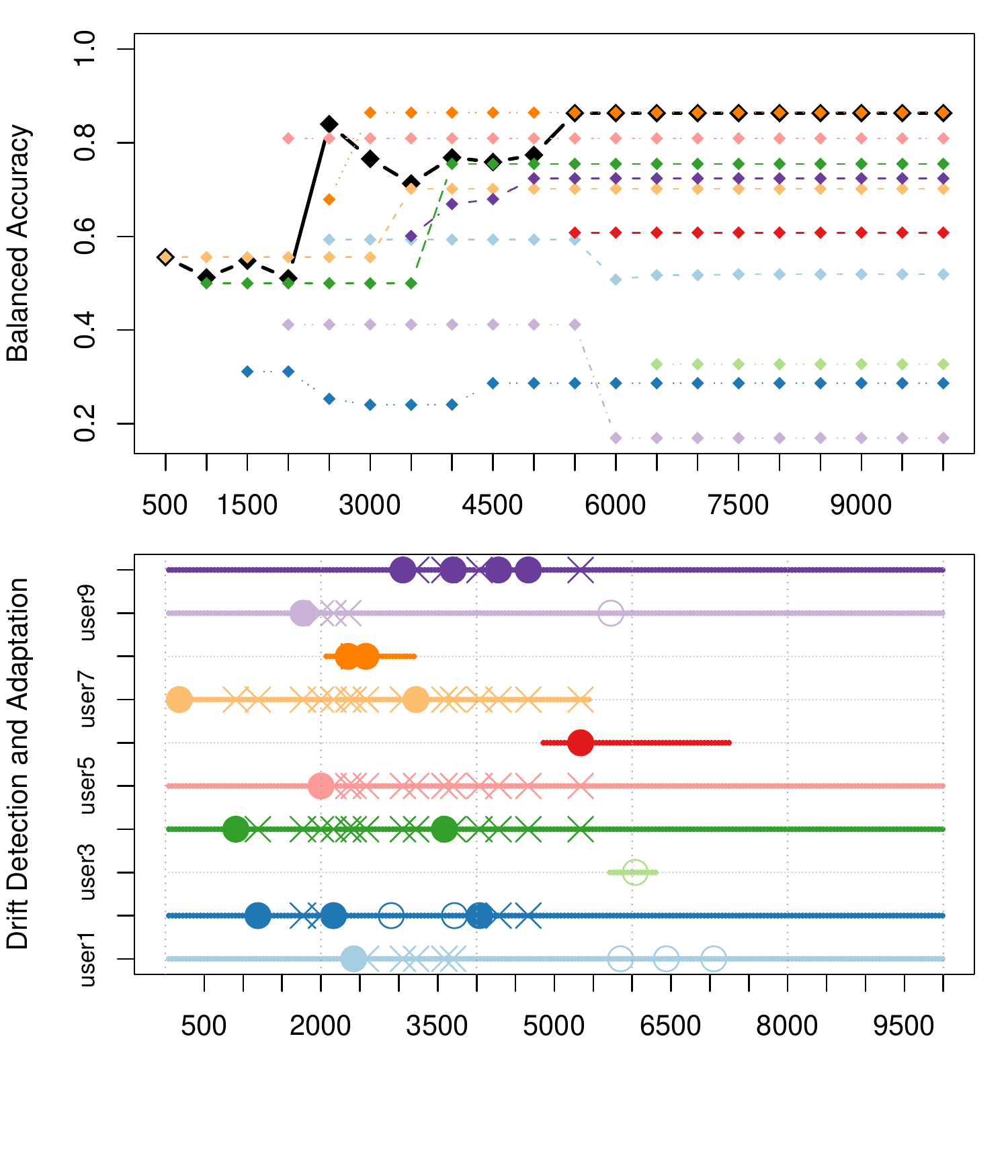}
\caption{Results for \ourAcronym{} (using RBF SVM as base classifier and $M_g = M_l = 5$) when 4 of the users are mislabeling the data.} 
\label{fig:results-method-4noisyusers}
\end{figure}

As we can see, \ourAcronym{} allows the society of devices to have a classifier from practically the beginning and to evolve it over time. In order to illustrate some other advantages of our approach, we have artificially modified the training dataset in different ways. Suppose now there are some users who are mislabeling the data, whether intentionally or not. To simulate this, we have inverted all the labels of four of the users. This is the maximum number of local datasets we can poison so that the system continues to properly choose the best participants for the global ensemble. In this example we have chosen three very active users (users 1, 2 and 9), and one more not very involved (user 9). The results using a single classifier trained on all the available data are those in Table~\ref{tab:results-old-4noisyusers}. The results using our method with different base classifiers are those shown in Table~\ref{tab:results-method-4noisyusers}. Figure~\ref{fig:roc-4nu} compares the ROC curves of the traditional models with the federated ones in this scenario. As can be appreciated, in this case \ourAcronym{} far outperforms the results provided by the classical approach. This is because the system is able to identify those noisy users and remove them from the global ensemble. In addition, the devices of those users will use the global model to label the unlabeled data correctly, thus overcoming the data that was manually mislabeled. Figure~\ref{fig:results-method-4noisyusers} shows the details of our federated and continual learning process in this scenario, in the particular case where RBF SVM is used as base classifier.

\begin{table}[htb]
\centering
\caption{Performance of traditional classifiers when there are users who provide noisy data.}
\label{tab:results-old-4noisyusers}
\resizebox{\linewidth}{!}{%
\begin{tabular}{l|c|c|c|}
	\cline{2-4}
	& \multicolumn{1}{c|}{\textbf{Balanced Accuracy}} & \multicolumn{1}{c|}{\textbf{Sensitivity}} & \multicolumn{1}{c|}{\textbf{Specificity}} \\ \hline
	\multicolumn{1}{|l|}{\textbf{Na{\"\i}ve Bayes \quad}} & 0.4837 & 0.2008 & 0.7667 \\ \hline
	\multicolumn{1}{|l|}{\textbf{GLM}} & 0.6283  & 0.5321 & 0.7245 \\ \hline
	\multicolumn{1}{|l|}{\textbf{C5.0}} & 0.4293 & 0.4645 & 0.3941 \\ \hline
	\multicolumn{1}{|l|}{\textbf{RBF SVM}} & 0.5943 & 0.6986 & 0.4899 \\ \hline
	\multicolumn{1}{|l|}{\textbf{RF}}      & 0.4679 & 0.5102 & 0.4256 \\ \hline
	\multicolumn{1}{|l|}{\textbf{SGB}} & 0.4499  & 0.3916 & 0.5082  \\ \hline
\end{tabular}%
}
\end{table}

\begin{table}[htbp]
\centering
\caption{Performance of \ourAcronym{} when there are users who provide noisy data ($M_g = M_l = 5$).}
\label{tab:results-method-4noisyusers}
\resizebox{\linewidth}{!}{%
\begin{tabular}{l|c|c|c|}
	\cline{2-4}
	& \multicolumn{1}{c|}{\textbf{Bal. Accuracy}} & \multicolumn{1}{c|}{\textbf{Sensitivity}} & \multicolumn{1}{c|}{\textbf{Specificity}} \\ \hline
	\multicolumn{1}{|l|}{\textbf{\ourAcronym{} + NB}}  & 0.7572 & 0.6116 & 0.9029 \\ \hline
	\multicolumn{1}{|l|}{\textbf{\ourAcronym{} + GLM}}  & 0.7495 & 0.9940 & 0.5050 \\ \hline
	\multicolumn{1}{|l|}{\textbf{\ourAcronym{} + C5.0}}  & 0.7834 & 0.8681 & 0.6986 \\ \hline
	\multicolumn{1}{|l|}{\textbf{\ourAcronym{} + SVM}}  & 0.8632 & 0.8356 & 0.8909 \\ \hline
	\multicolumn{1}{|l|}{\textbf{\ourAcronym{} + RF}}  & 0.8137 & 0.9888 & 0.6387 \\ \hline
	\multicolumn{1}{|l|}{\textbf{\ourAcronym{} + SGB}} & 0.8269  & 0.9684 & 0.6854 \\ \hline
\end{tabular}%
}
\end{table}

\begin{figure}[htb]
\centering
\includegraphics[width=0.9\linewidth,trim={0 2cm 0 0},clip]{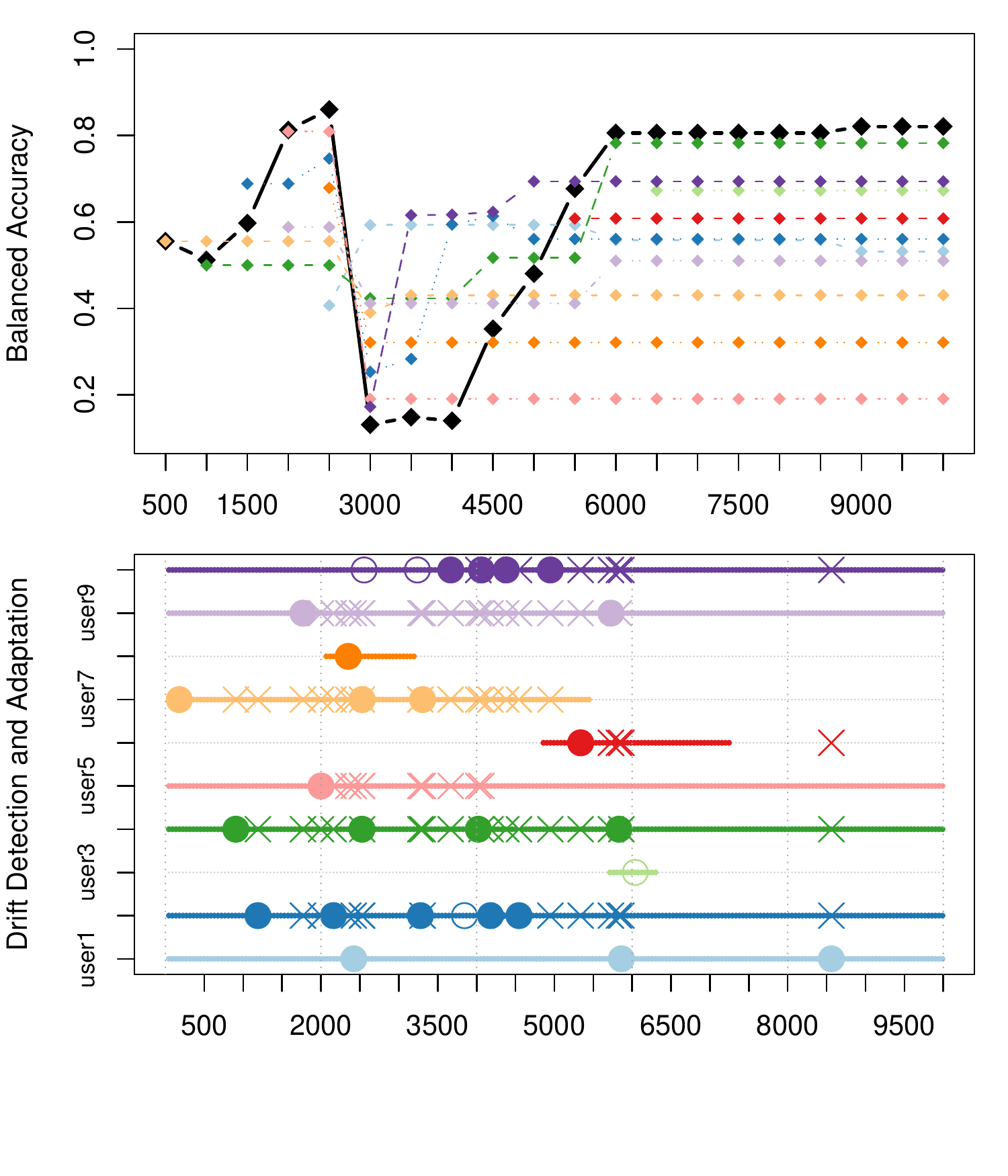}
\caption{Results for \ourAcronym{} (using RBF SVM as base classifier and $M_g = M_l = 5$) when a drastic drift occurs at $t=2500$.} 
\label{fig:results-method-totaldrift}
\end{figure}

Finally, we have simulated a big drift in data. The biggest drift the system could experience would be a total inversion in the meaning of the two classes. Therefore, from the instant $t=2500$, all devices begin to identify patterns of ``walking'' when before they would be ``not walking'', and vice versa. Figure~\ref{fig:results-method-totaldrift} shows the process using again the RBF SVM as base classifier. As we can see, it takes a while for the system to converge because the change is too drastic, but it ends up achieving it. Obviously, if we reduce the size of the local ensembles, $M_l$, the system would adapt faster.

\section{Conclusions}
\label{sec:conclusions}
In this paper we have presented \ourAcronym{}, a novel method for light, semi-supervised, asynchronous, federated, and continual classification. Our method allows to face real tasks where several distributed devices (smartphones, robots, etc.) continuously acquire biased and partially-labeled data, which evolves over time in unforeseen ways. It basically consists of training local models in each of the devices and then share those models with the cloud, where a global model is created. This global model is sent back to the devices and helps them to label unlabeled data to improve the performance of their local models. We have applied our proposal to a real classification task, walking recognition, and we have shown that our proposal obtains very high performances without the need of large amounts of labeled data collected at the beginning. 

We believe that we have opened a very promising line of research in the context of federated learning, and that there is still a lot of work to be done. In this work we have presented a new method with clearly separated components: local learning, semi-supervised labeling, local drift detection and adaptation, and selection of local candidates for the subsequent global consensus. However, we believe that all of these modules can be further improved. For example, we want to explore optimal ways to combine models locally and also in the cloud, study effective ways to perform distributed feature selection, analyze the use of the global model for instance selection in the local devices or the application of techniques such as amending~\cite{triguero2015}.
On the other hand, real world, physically distributed devices have an intrinsic data skewness property so, depending on the problem, instead of having a single global model, it could be more interesting to cluster users or devices that share similar properties and then obtain several global models, one for each group. Another promising approach could be to have some system of local adaptation of the global model, for example using transfer learning, so that it would be able to adjust to local particularities.

\section*{Acknowledgements}
This research has received financial support from AEI/FEDER (European Union) grant number TIN2017-90135-R, as well as the \textit{Conseller\'ia de Cultura, Educaci\'on e Ordenaci\'on Universitaria} of Galicia (accreditation 2016--2019, ED431G/01 and ED431G/08, and reference competitive group ED431C2018/29), and the European Regional Development Fund (ERDF). It has also been supported by the \textit{Ministerio de Educaci\'on, Cultura y Deporte} of Spain in the FPU 2017 program (FPU17/04154).

\section*{Declarations of interest}
None.


\bibliographystyle{elsarticle-num} 
\bibliography{main}

\end{document}